\documentclass[journal]{IEEEtran}
\usepackage{floatrow}
\newfloatcommand{capbtabbox}{table}[][\FBwidth]

\usepackage{cite}
\usepackage{pdfpages}
\usepackage{amsmath}
\usepackage{diagbox}
\usepackage{tabularx} 
\usepackage{multirow}
\usepackage{multicol}
\usepackage{arydshln}
\usepackage{makecell}
\usepackage{booktabs}
\usepackage{amssymb}
\usepackage{color}
\usepackage[colorlinks,linkcolor=black,anchorcolor=black,citecolor=black,urlcolor = black
]{hyperref}

\usepackage{array}
\usepackage[caption=false,font=normalsize,labelfont=sf,textfont=sf]{subfig}
\usepackage{textcomp}
\usepackage{stfloats}
\usepackage{url}
\usepackage{verbatim}
\usepackage{graphicx}
\usepackage{booktabs} 

\begin{document}
\newcommand{\bb}[1]{\mathbf{#1}}
\newcommand{\bma}{\bb{a}}
\newcommand{\bmb}{\bb{b}}
\newcommand{\bmc}{\bb{c}}
\newcommand{\bmd}{\bb{d}}
\newcommand{\bmf}{\bb{f}}
\newcommand{\bmg}{\bb{g}}
\newcommand{\bmh}{\bb{h}}
\newcommand{\bmr}{\bb{r}}
\newcommand{\bms}{\bb{s}}
\newcommand{\bmt}{\bb{t}}
\newcommand{\bmu}{\bb{u}}
\newcommand{\bmn}{\bb{n}}
\newcommand{\bmv}{\bb{v}}
\newcommand{\bmw}{\bb{w}}
\newcommand{\bmx}{\bb{x}}
\newcommand{\bmy}{\bb{y}}
\newcommand{\bmz}{\bb{z}}

\newcommand{\bmA}{\bb{A}}
\newcommand{\bmO}{\bb{O}}
\newcommand{\bmB}{\bb{B}}
\newcommand{\bmC}{\bb{C}}
\newcommand{\bmD}{\bb{D}}
\newcommand{\bmF}{\bb{F}}
\newcommand{\bmG}{\bb{G}}
\newcommand{\bmH}{\bb{H}}
\newcommand{\bmI}{\bb{I}}
\newcommand{\bmL}{\bb{L}}
\newcommand{\bmS}{\bb{S}}
\newcommand{\bmM}{\bb{M}}
\newcommand{\bmN}{\bb{N}}
\newcommand{\bmR}{\bb{R}}
\newcommand{\bmT}{\bb{T}}
\newcommand{\bmU}{\bb{U}}
\newcommand{\bmV}{\bb{V}}
\newcommand{\bmW}{\bb{W}}
\newcommand{\bmX}{\bb{X}}
\newcommand{\bmY}{\bb{Y}}
\newcommand{\bmZ}{\bb{Z}}
\newcommand{\bmK}{\bb{K}}
\newcommand{\bmp}{\partial}

\newcommand{\bmin}{\bb{in}}
\newcommand{\bmout}{\bb{out}}
\newcommand{\bmSCR}{\bb{SCR}}
\newcommand{\bmSCRG}{\bb{SCRG}}
\newcommand{\bmBSF}{\bb{BSF}}
\newcommand{\bmCG}{\bb{CG}}
\newcommand{\bmCON}{\bb{CON}}
\newcommand{\bmPd}{\bb{Pd}}
\newcommand{\bmFa}{\bb{Fa}}

\newcommand{\bmtrue}{\bb{true}}
\newcommand{\bmtargets}{\bb{targets}}
\newcommand{\bmdetected}{\bb{detected}}
\newcommand{\bmtotal}{\bb{total}}
\newcommand{\bmfalse}{\bb{false}}
\newcommand{\bmpixels}{\bb{pixels}}
\newcommand{\bmimages}{\bb{images}}

\newcommand{\bmzeta}{\mbox{\boldmath $\zeta$}}
\newcommand{\bmPhi}{{\boldsymbol{\Phi}}}
\newcommand{\bmeta}{\bm{\mathrm{\eta}}}
\newcommand{\bmxi}{\bm{\mathrm{\xi}}}

%
\ifCLASSINFOpdf
\else
\fi
\hyphenation{op-tical net-works semi-conduc-tor}

%
\title{Infrared UAV Target Tracking with Dynamic Feature Refinement and Global Contextual Attention Knowledge Distillation}




%
%
%

\author{\thanks{Manuscript received xx xx, 2025.}
	\thanks{This work was supported in part by the Open Research Fund of the National Key Laboratory of Multispectral Information Intelligent Processing Technology under Grant 61421132301, the National Natural Science Foundation of China under Grants 61971460 and 62101294.  \emph{(Corresponding author: Kun Bai; Houzhang Fang}.)}
	Houzhang~Fang,~\IEEEmembership{Member,~IEEE}, 
	Chenxing~Wu,
		Kun~Bai,
        Tianqi~Chen,
       Xiaolin~Wang,
  Xiyang~Liu,
Yi~Chang,~\IEEEmembership{Member,~IEEE},
        Luxin~Yan,~\IEEEmembership{Member,~IEEE} 

\thanks{H. Fang, C. Wu, T. Chen, X. Wang, and X. Liu are with the School of Computer Science and Technology, Xidian University, Xi'an 710126, China (e-mail: houzhangfang@xidian.edu.cn, wcx1111111029@163.com, 18957384576@163.com, wxl@stu.xidian.edu.cn, xyliu@xidian.edu.cn)}
\thanks{K. Bai is with the Xi’an Modern Control Technology Research Institute, Xi'an 710065, China (baikundb@126.com)}
\thanks{Y. Chang and L. Yan are with the School of Artificial Intelligence and Automation, Huazhong University of Science and Technology, Wuhan 430074, China (yichang@hust.edu.cn, yanluxin@hust.edu.cn).}

}

%
%

\markboth{Journal of \LaTeX\ Class Files,~Vol.~13, No.~9, September~2014}%
{Shell \MakeLowercase{\textit{et al.}}: Bare Demo of IEEEtran.cls for Journals}
%



\maketitle

\begin{abstract}
Unmanned aerial vehicle (UAV) target tracking based on thermal infrared imaging has been one of the most important sensing technologies in anti-UAV applications. However, the infrared UAV targets often exhibit weak features and complex backgrounds, posing significant challenges to accurate tracking. To address these problems, we introduce SiamDFF, a novel dynamic feature fusion Siamese network that integrates feature enhancement and global contextual attention knowledge distillation for infrared UAV target (IRUT) tracking. 
The SiamDFF incorporates a selective target enhancement network (STEN), a dynamic spatial feature aggregation module (DSFAM), and a dynamic channel feature aggregation module (DCFAM). The STEN employs intensity-aware multi-head cross-attention to adaptively enhance important regions for both template and search branches.
The DSFAM enhances multi-scale UAV target features by integrating local details with global features, utilizing spatial attention guidance within the search frame.
The DCFAM effectively integrates the mixed template generated from STEN in the template branch and original template, avoiding excessive background interference with the template and thereby enhancing the emphasis on UAV target region features within the search frame.
Furthermore, to enhance the feature extraction capabilities of the network for IRUT without adding extra computational burden, we propose a novel tracking-specific target-aware contextual attention knowledge distiller (TCAKD). It transfers the target prior from the teacher network to the student model, significantly improving the student network's focus on informative regions at each hierarchical level of the backbone network.
Extensive experiments on real infrared UAV datasets demonstrate that the proposed approach outperforms state-of-the-art target trackers under complex backgrounds while achieving a real-time tracking speed.
\end{abstract}

\begin{IEEEkeywords}
UAV surveillance, infrared target tracking, feature fusion, knowledge distillation, attention mechanism.
\end{IEEEkeywords}

%
\IEEEpeerreviewmaketitle

\section{Introduction}\label{section1}
\begin{figure}[!t]
\centering 
\includegraphics[width=3.5in,keepaspectratio]{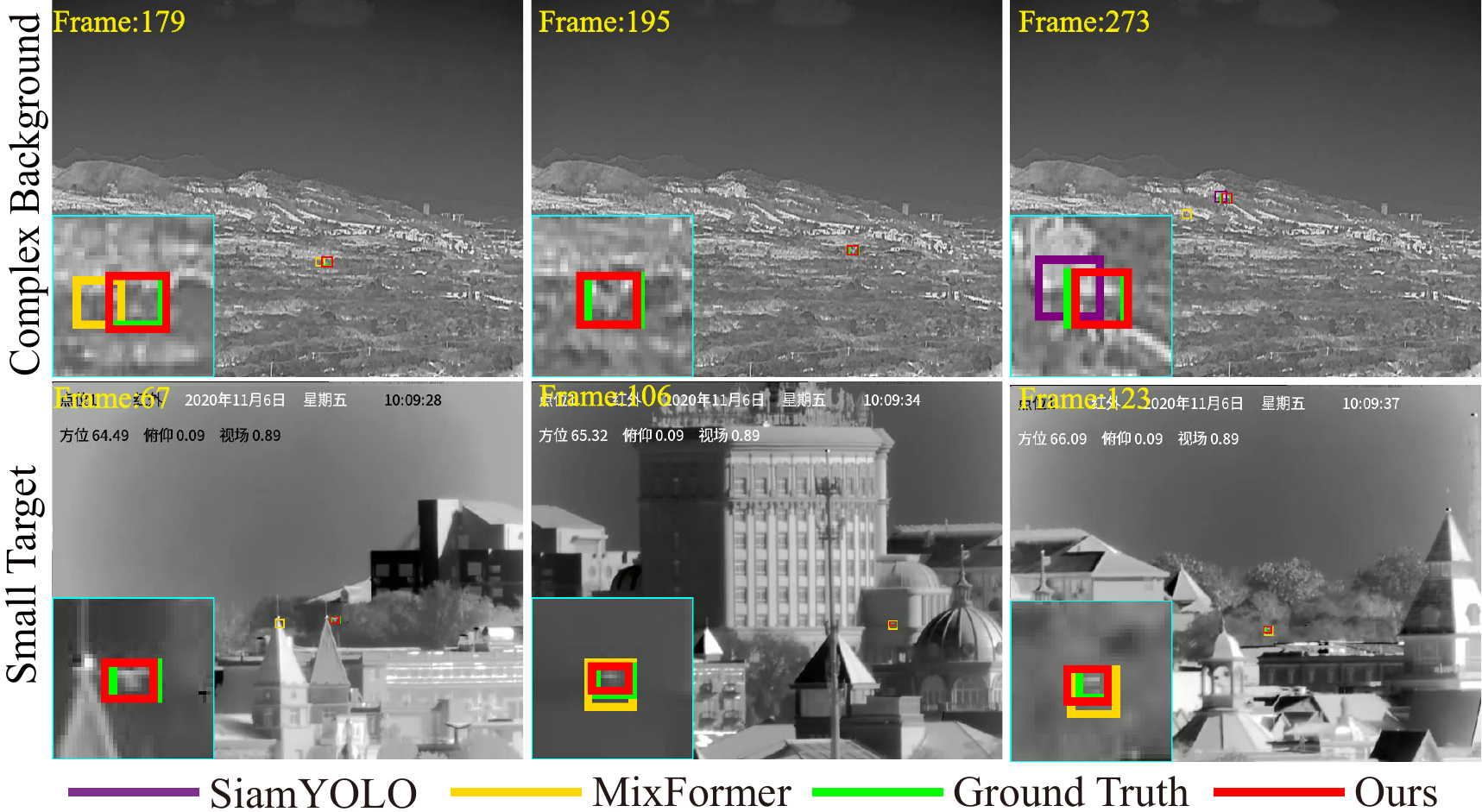}
\caption{Qualitative comparison of our method with the baseline SiamYOLO \cite{2021ICCVWFang} and MixFormer \cite{cuimixformer} on two challenging video sequences (complex background and small target). The proposed Siamese tracker demonstrates superior performance in  infrared UAV target tracking across various complex backgrounds due to its dynamic feature enhancement and global contextual attention-based knowledge distillation.}
\label{fig1} 
\end{figure}
\IEEEPARstart{R}{ecently}, unmanned aerial vehicles (UAVs) have been widely applied in many fields, such as precision agriculture, environmental monitoring, and aerial photography \cite{2025CVPR_UniCD,2022TIMFang,2023TII_DAGNet,2023ACMMM_DANet,fang2024online}. However, the wide use of UAVs can also potentially pose great threats to both aerial safety and public security. UAV surveillance based on infrared thermal imaging is capable of monitoring UAVs at a long range in both day and night scenarios \cite{2025CVPR_UniCD,2022TIMFang,2023TII_DAGNet,2023ACMMM_DANet}. As a key perception technology for UAV surveillance in the anti-UAV system, infrared UAV target (IRUT) tracking is a very challenging problem due to weak target features, small in scale, and distractors in complex backgrounds.

Many works have been developed for target tracking task. SiamFC \cite{2016ECCVWBertinetto} initially introduced deep learning into the target tracking task and utilized a correlation operation to obtain a similarity matrix for locating the target. This method is further enhanced by SiamRPN \cite{2018CVPRSiamRPN} and SiamRPN++ \cite{2019CVPRSiamRPN++}. However, these methods are susceptible to distractors since the linear correlation operations between template and search frame. Recently, Transformer has been successfully applied in target tracking fields \cite{2021_ICCV_Stark,cai2023robust,YuDomain,zheng2024odtrack,cuimixformer} and significantly improved the tracking performance. However, these methods utilize all pair-wise attention of Transformer, lacking a focus on the primary areas. The IRUT size is relatively small, leading to a reduced importance weight for informative areas. Meanwhile, the cross-attention introduces excessive background into the template frame, which undermines the features of small IRUT. This makes the subsequent enhancement of target features in the search frame more susceptible to interference from similar objects.

For the IRUT tracking task, Huang et al. \cite{huang2023searching} employed a C-C RPN to coarsely select candidate regions, which are combined with reliable historical predictions and sent into an S-L RCNN for accurate prediction. Fang et al. \cite{fang2024online} constructed a contrast-enhanced block and a full spatial resolution attention mechanism to improve the representation ability of IRUT. Yu et al. \cite{Yu_2023_CVPR} proposed a local-global tracker that employs a Transformer-based approach as the local tracker and utilizes a multi-region local tracking module to track potential target regions simultaneously. However, these methods fail to capture the local details of IRUT, which makes it difficult to highlight multi-scale target features in the search frames under complex scenarios.

Accurate and efficient target tracking is essential for practical applications. Some work \cite{chen2022efficient,chen2022backbone,kang2023exploring,gopal2024separable} has been developed to address the trade-off between speed and performance. However, the performance still has a big gap compared with networks with larger parameter sizes. Knowledge distillation \cite{hinton2015distilling} is viewed as an effective method for enhancing the performance of lightweight models without affecting inference efficiency.
Shen et al. \cite{shen2021distilled} were the first to introduce knowledge distillation into Siamese trackers, designing a Siamese target response to effectively transfer tracking-specific knowledge from the teacher model to the student model. However, due to the small scale of the IRUT, the calculation of the Siamese target response relies on a local correlation process, which can easily result in a local optimum and is particularly susceptible to distractors. This leads to inaccurate knowledge transfer.

In this paper, we present a novel dynamic feature fusion Siamese network (SiamDFF) for infrared UAV target tracking. Specifically, we first introduce a selective target enhancement network (STEN) that integrates intensity-aware multi-head cross-attention to focus on the most relevant information in all pairwise relations and reduce the influence of excessive background information on the learning of correlation attention weights between pairs of pixels, thereby effectively enhancing the features of the target regions. 

We then propose a dynamic spatial feature aggregation module (DSFAM) to improve the UAV target feature representation capability in the search frame. DSFAM employs a local-global complementary dual-branch structure, where the CNN-based branch effectively extracts local spatial details, and the Transformer-based branch captures global contextual information. After processing through STEN, the Transformer-based features provide a coarse representation of the target's shape. When combined with the CNN-based features, this allows for a complementary enhancement of target details, making them more focused and prominent. To adaptively utilize representations from both types of features, DSFAM facilitates feature interaction between the two branches, generating two spatial attention maps. These attention maps are then applied to each respective branch, enhancing the target-specific features across both branches.
Finally, we introduce a dynamic channel feature aggregation module (DCFAM) to simultaneously integrate the original template and the mixed template generated from STEN in the template branch. DCFAM effectively prevents excessive contamination of the template by background elements. The dual-branch structure promotes target-related discriminative feature learning, enhancing mutual interaction with the search frame features and improving the representation of target region features within the search frame. DCFAM facilitates bidirectional feature interaction between the two types of templates, generating channel-wise attention maps through local cross-channel interactions and multiplying with the different branches to recalibrates and enhances the corresponding branches. 

Furthermore, to enhance the feature extraction capability of the backbone without introducing additional computational burden, we propose the target-aware contextual attention knowledge distiller (TCAKD). TCAKD initially utilizes prior knowledge of the target mask from the template frame to focus on the relevant target features. It then employs a Transformer-based architecture to model long-range dependencies between the template and the search frame, resulting in a tracking-specific attention map that highlights the target region while suppressing background interference. The teacher model can better extract IRUT features, which improves the precision of template matching process and leads to a high response to the target region features. This refined knowledge is then transferred to the student network, assisting it in learning the template matching patterns established by the teacher network. The qualitative comparison of our method with the other two baseline trackers is shown in Fig. \ref{fig1}. 

The main contributions of this article are summarized as follows. 
\begin{itemize}
\item We propose a novel framework SiamDFF for consistently and accurately tracking IRUTs in complex backgrounds. It utilizes STEN to dynamically adjust the weights of pairwise relations in the Transformer to enhance the informativeness of the feature representation.
We also introduce a DSFAM to effectively combine local detail information and global contextual information of IRUT to improve the feature representation ability in the search frame.
Furthermore, to mitigate the template contamination issue, we design a DCFAM that utilizes the inherent information of the target template to complement the target details.

\item We design a knowledge distiller TCAKD to enhance tracking performance while maintaining high inference speed by transferring a target-focused spatial attention map to the student model to learn the matching patterns derived from the teacher model.

\item We conduct a comprehensive comparison of various tracking methods on real-world IRUT dataset. Extensive experiments demonstrate that our method
achieves superior tracking performance against other SOTA methods, and realizes real-time tracking.
\end{itemize}

\section{Related Work}  \label{section2}
\subsection{Infrared Target Tracking Methods}
In recent years, infrared target tracking has gained increasing attention. Liu et al. \cite{2021TMMLiu} introduced a dual-level feature model that integrates discriminative features with fine-gained correlation features to enhance infrared object representation. Liu et al. \cite{2021TMMLiu} proposed a multi-level similarity model combined with global semantic similarity and local structural similarity for infrared object tracking. Fang et al. \cite{2021ICCVWFang} developed an infrared UAV tracker based on global search, employing spatial-temporal information to counter distractions and using attention mechanisms to fuse features from different levels, thus enhancing target feature representation. Jiang et al. \cite{2021TMMJiang} constructed a large-scale anti-UAV dataset and proposed a dual-flow semantic consistency training strategy to differentiate tracked instances from distractors. Another class of long-term tracking methods combines global detection  \cite{2022CVPRISNet,2022ACMMMRKformer,2023TGRSDim2Clear,2023TII_DAGNet,2024ECCVIRSAM} with local tracking, such as TOPIC \cite{2025TIPTOPIC}.  The detection methods \cite{2022CVPRISNet,2022ACMMMRKformer,2023TGRSDim2Clear,2023TII_DAGNet,2024ECCVIRSAM} have demonstrated strong effectiveness in detecting infrared small targets. This type of tracking method performs well on targets in simple scenarios, but still face challenges when dealing with small-scale IRUTs in complex backgrounds and under real-time processing requirements. Despite recent advancements, the small size and weak features of IRUTs continue to pose significant challenges for tracking methods, making it difficult to effectively extract discriminative features and limiting their salient representation in the search frame. While these methods can track infrared targets, they tend to overlook the feature interaction phase between template and search frame, which is crucial for enhancing target features in the search frame. Additionally, there is insufficient focus on improving the feature extraction capabilities of the backbone, which impacts the subsequent template matching process and leads to inaccuracies in tracking. In light of this, we focus on enhancing the feature interaction process and improving the feature extraction capability for IRUTs.

\begin{figure*}[!thbp]
\centering
\setlength{\abovecaptionskip}{-0.1em}
\includegraphics[width=\linewidth,keepaspectratio]{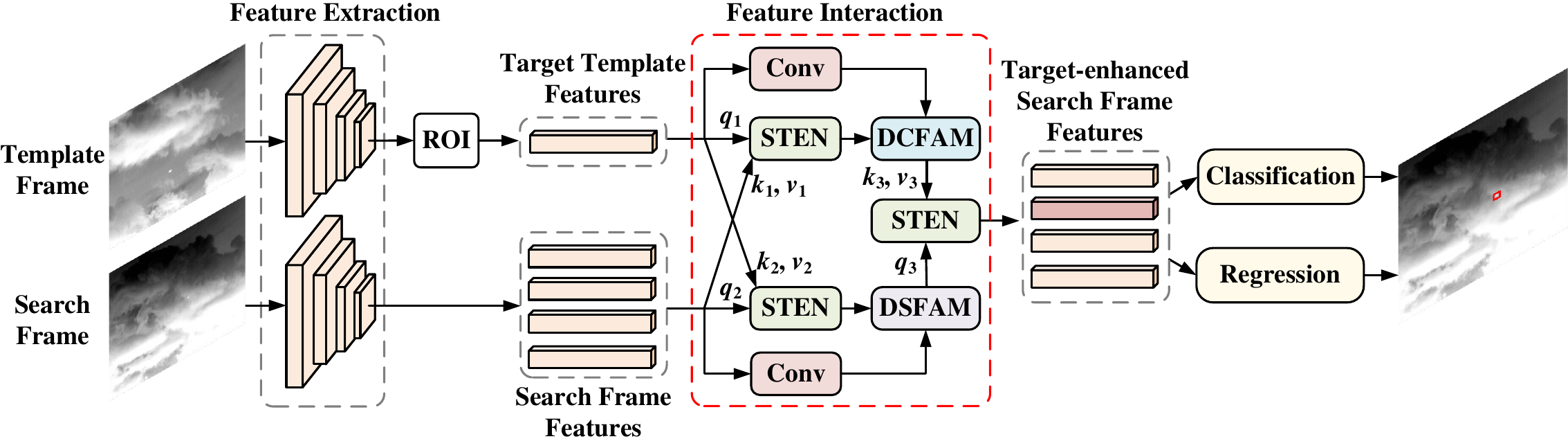}
\caption{Overview of the proposed SiamDFF. The feature interaction module enhances search frame features to improve the ability of the model for target classification and localization, which includes the proposed STEN, DSFAM, and DCFAM. The ROI denotes region of interest. 
}
\label{siamdff}
\end{figure*}

\subsection{Knowledge Distillation for Target Tracking}  
Knowledge distillation is a model compression technique that improves model performance by transferring knowledge from a teacher network to a lightweight student network. For target tracking task, Wang et al. \cite{wang2020real} developed a fidelity loss and a correlation tracking loss to effectively compress the network while transferring its focus from object recognition to visual tracking. Shen et al. \cite{shen2021distilled} proposed a Siamese target response learning algorithm that facilitates the transfer of tracking-specific knowledge to the student network. However, these distillation strategies focus solely on local matching information related to tracking-specific loss, neglecting contextual feature information. Consequently, distractor information is included in the feature response maps transferred to the student network, which hinders its ability to effectively learn the target region. In this study, we propose the TCAKD, which transfers the global contextual attention generated by the teacher model to the student network, enhancing the ability of the student network to perceive the target during feature extraction and improving its robustness against interference.

\section{Methodology} \label{section3}

\subsection{SiamDFF Framework} \label{subsection_CMORB}
In this subsection, we introduce the proposed SiamDFF framework, as depicted in Fig. \ref{siamdff}. We employ ResNet18 \cite{He_2016_CVPR} as the backbone network for feature extraction, ensuring the efficiency of our framework. 
The feature interaction module comprises STEN, DSFAM, and DCFAM for enhancing the representation capability of the target feature within the search frame. Finally, the target-enhanced search frame features are sent to the anchor-free head for the final prediction.
\subsubsection{Selective Target Enhancement Network}


The feature interaction process is a crucial step in the Siamese network for target tracking. 
Transformer-based interaction has become a mainstream paradigm in current tracking framework due to its ability to capture long-range dependencies and integrate contextual information.
However, the cross-attention performs a one-to-all spatial similarity calculation for feature aggregation, which can introduce redundant information and hinder the ability of the model to establish accurate long-range dependencies.
For small IRUTs occupying only a small portion of the image, dependence on pairwise similarity increases the weights of the background, resulting in inaccurate enhancement of the target features.
Existing works \cite{gao2022aiatrack,ijcai2022p127} aim to adjust the correlation computation weight map in the self-attention module to reduce background interference. However, their modules require substantial computational resources, significantly reducing the  efficiency of the model.

To achieve accurate and focused feature interaction in cross-attention for IRUT, we propose STEN, which combines intensity-aware multi-head cross-attention (IMC) that aggregates the most useful global context to each pixel. IMC first calculates all pairwise relations between the query and key to form a similarity map. The aggregated global context for each query focus varies at different positions within an image. Therefore, we establish a threshold that enables dynamic adjustment. Specifically, for each pixel in feature map, we utilize the values of the top k percent of positions based on their similarity to all other pixels to identify which regions should be considered for feature aggregation. As a result, enhancing the ability of the model to exploit global features while primarily focusing on the most relevant areas. 

\begin{figure}[!t]
\centering
\setlength{\abovecaptionskip}{-0.05em}
\includegraphics[width=3in,keepaspectratio]{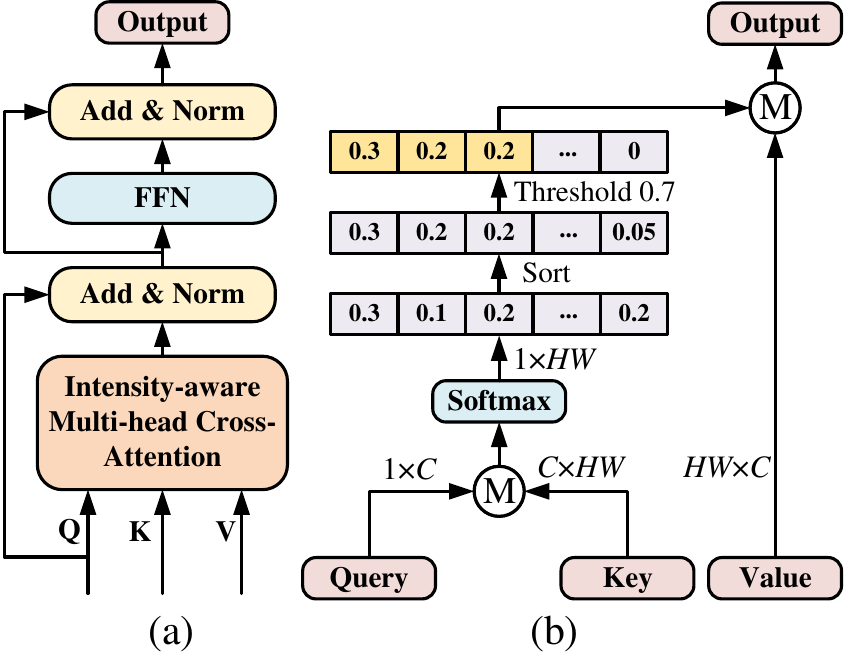}
\caption{(a) Structure of selective target enhancement network (STEN). (b) Structure of intensity-aware multi-head
cross-attention (IMC). $\raisebox{0.1pt}{\textcircled{\raisebox{-0.1pt}{\tiny M}}}$ denotes the matrix multiplication.}
\label{sten}
\end{figure}

The constructed STEN is shown in Fig. \ref{sten}(a). We first send Q, K, and V into IMC to precisely model the relation between the template and search frame feature. The calculation of STEN can be formulated as follows:
\begin{align}\label{sten}
X&=\operatorname{Norm}\left(\operatorname{IMC}\left(\boldsymbol{Q}, \boldsymbol{K}, \boldsymbol{V}\right)+\boldsymbol{Q}\right), \nonumber \\
Y&=\operatorname{Norm}\left(\operatorname{FFN}\left(\boldsymbol{X}\right)+\boldsymbol{X}\right),
\end{align}
where FFN and Norm denote the feed forward network and layer normalization, respectively. $X$ and $Y$ denote the outputs from IMC and STEN.

The IMC is shown in Fig. \ref{sten}(b). We begin by computing the all pairwise relations between the query and key. Next, we sort the values in the similarity matrix in descending order. We then sequentially retain the top-ranked larger values until the cumulative sum of the selected values reaches the specified threshold $T$, ensuring that we retain the most informative elements. We compute the cumulative sum sequentially from the first element.  For low attention weight elements, we set their values to zero. The calculation of IMC can be formulated as follows:
\begin{align}\label{sten}
\operatorname{IMC}&=\mathrm{softmax}\left(\mathcal{F}\left(\frac{\mathbf{Q}\mathbf{K}^{\top}}{{\sqrt{d}}}\right)\right)\mathbf{V},
\end{align}
where $d$ represents the dimension of the key. $\mathcal{F}$ is a function that determines which elements to discard:
\begin{align}\label{sten}
[\mathcal{F}(\boldsymbol{E})]_{ij} = \begin{cases} 
E_{ij}, & \text{if } S_{ij} \geq t_i, \\
0, & \text{otherwise.}
\end{cases}
\end{align}
where $t_i$ is the smallest element that enables the cumulative sum to reach the specified threshold $T$ in the $i$-th row of $\frac{\mathbf{Q}\mathbf{K}^{\top}}{{\sqrt{d}}}$. The selection of the  specified threshold $T$ is presented in Table \ref{sten_thres} of the ablation study.

In the search branch, after passing through the STEN module, irrelevant context is effectively discarded, thereby enhancing the search frame's ability to perceive the target. Meanwhile, for template branch, STEN improves the template frame's discriminative capability between the target and background, facilitating the decoder's further emphasis on the features of the target region.
\subsubsection{Dynamic Spatial Feature Aggregation Module}
\begin{figure}[!t]
\centering
\setlength{\abovecaptionskip}{-0.1em}
\includegraphics[width=3.2in,keepaspectratio]{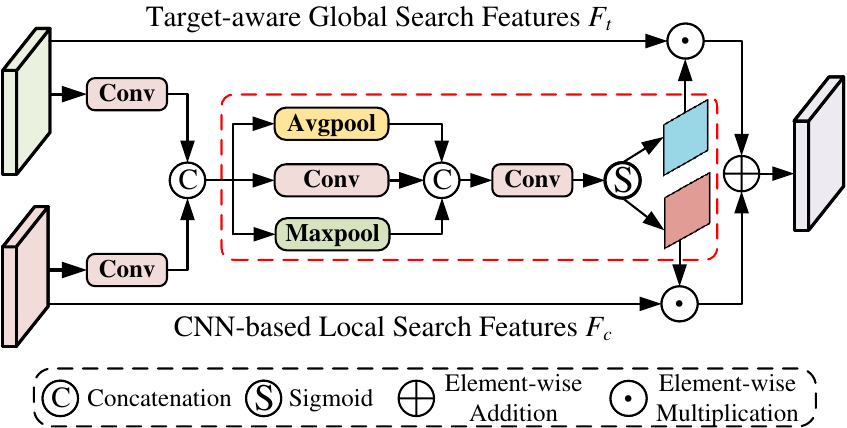}
\caption{Architecture of the proposed DSFAM.}
\label{dsfam}
\end{figure}
The Transformer-based feature interaction effectively leverages global contextual information to model dependencies and encodes spatial relationships, enhancing the robustness of the model against interference and assisting in capturing the shape details of targets across various scales. However, global features often lack the local detail necessary for precise IRUT representation. Meanwhile, the global perspective of Transformers is advantageous for detecting large-scale targets, their ability to extract features from small-scale targets is limited \cite{zhu2020deformable}. The local feature can provide fine-gained details essential for accurate target tracking. By effectively combining the strengths of both, we can significantly enhance the network's capability to track multi-scale IRUTs.

We propose a DSFAM, which aggregates Transformer-based features and CNN-based features using a spatial attention mechanism. Simply adding or concatenating the two branches leads to insufficient correlation between the features. Instead, we fuse their features and apply a transformation to generate spatial attention weight maps that contain target-aware spatial information. DSFAM allows the model to dynamically adjust the weights of each branch, precisely enhancing the target representation and resulting in a more effective fusion way. Unlike convolution operations that indiscriminately extract features of a certain type without considering the specific target, DSFAM leverages the target perception capabilities of the Transformer branch while selectively utilizing the relevant features extracted by the CNN branch. By employing an attention-guided fusion strategy, DSFAM achieves a more fine-gained integration of features, maximizing the complementary advantages of both branches and improving the overall performance in target tracking.

As shown in Fig \ref{dsfam}, we first element-wise add the features from both branches. We then employ spatial attention maps to determine the importance of each branch. Given the input target-aware global search feature is ${F_t}\in\mathbb{R}^{{C}\times{H}\times{W}}$ and CNN-based local feature is ${F_c}\in\mathbb{R}^{{C}\times{H}\times{W}}$. We first squeeze their channel to $\frac{C}{2}$ by ${1}\times{1}$ convolution to reduce the computational burden of subsequent operations. Then we concatenate the output feature and obtain the fused feature denoted as ${F_f}\in\mathbb{R}^{{C}\times{H}\times{W}}$. The output of ${F_f}$ can be formulated as 
\begin{align}\label{fused feature}
F_{f}=\operatorname{Cat}\left(\operatorname{Conv_1}(F_{t}), \operatorname{Conv_1}(F_{c}) \right),
\end{align}
where $\operatorname{Conv_1}$ and Cat denote the $1\times1$ convolution and concatenation operation, respectively.

Next, we apply convolution, average pooling, and max pooling to obtain the attention response in the spatial dimension. The global pooling operations are employed to capture global spatial statistical characteristics, while the convolution component is used to integrate local spatial information. Then the three obtained feature maps concatenate in channel-wise and then utilize $3\times3$ convolution to obtain $w_t$ and $w_c$ which imply the spatial coefficients for each branch. Afterward, a sigmoid function is applied to separately emphasize the importance of spatial region. The formulation can be denoted as
\begin{align}\label{wc}
w_c=\operatorname{Conv_3}(F_{f}),
\end{align}
\begin{align}\label{wm}
w_{m}=\operatorname{MaxPool}(F_{f}),
\end{align}
\begin{align}\label{wa}
w_{a}=\operatorname{AvgPool}(F_{f}),
\end{align}
\begin{align}\label{ab}
\alpha_1,\beta_1=\sigma\left(\operatorname{Conv_3}\left( \operatorname{Cat}\left(w_c,w_m,w_a \right)\right)\right),
\end{align}
where $\operatorname{Conv_3}$ and $\sigma$ denote the $3\times3$ convolution and sigmoid function, respectively.  The final output on the right-hand side of Eq. (\ref{ab}) has dimensions of $H\times W\times2$, with the first channel corresponding to $\alpha_1$ and the second to $\beta_1$. 

Finally, the modulated feature maps $F_m$ are obtained by the Transformer-based feature and CNN-based feature based on the calculated $\alpha_1$ and $\beta_1$. The feature fusion operation can be denoted as
\begin{align}\label{fm}
F_m=\alpha_1*F_t + \beta_1*F_c.
\end{align}

\subsubsection{Dynamic Channel Feature Aggregation Module}
\begin{figure}[!t]
\centering
\setlength{\abovecaptionskip}{-0.1em}
\includegraphics[width=3.3in,keepaspectratio]{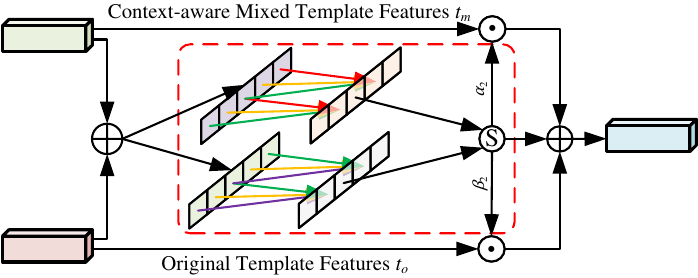}
\caption{Architecture of the proposed DCFAM.}
\label{dcfam}
\end{figure}
The quality of the template significantly influences tracking performance. Incorporating contextual information from the search frame into the template is beneficial, as it enables the template to adapt to variations in the target's shape and scale by leveraging the global context provided by the search frame. However, by interacting the template with the search frame, the mixed template incorporates a substantial amount of background area, which can lead to potential contamination by distractors \cite{cai2023robust}.

We propose DCFAM which prevents potential degradation issues while providing a discriminative target-oriented fused template feature. DCFAM employs channel attention maps to adjust the importance of the two type templates. Due to the typically small scale of UAV targets, directly compressing the channel dimension when calculating channel attention can inevitably lead to the loss of small target information. In DCFAM, we introduce channel attention without dimension compression \cite{2020CVPRWang}, allowing for a more accurate calculation of the importance weights for each branch.

As shown in Fig \ref{dcfam}, assume the mixed template is $t_m \in\mathbb{R}^{{C}\times{1}\times{1}}$ and origin template is $t_o \in\mathbb{R}^{{C}\times{1}\times{1}}$. We first fuse the two type templates and get fused template $t_f \in\mathbb{R}^{{C}\times{1}\times{1}}$. The process is denoted as
\begin{align}\label{tf}
t_f=t_m + t_o.
\end{align}
We then employ 1D convolution to obtain local cross-channel attention vectors for each branch. Afterward, a sigmoid function is applied to generate finally weight vectors. The procedure can be denoted as
\begin{align}\label{tf}
\alpha_2,\beta_2=\sigma\left(\operatorname{Conv1_m(t_m),Conv1_o(t_o)}\right),
\end{align}
where $\operatorname{Conv1}_m(t_m)$ and $\operatorname{Conv1}_o(t_o)$ denote 1D convolution used for mixed template and 1D convolution used for origin template, respectively. 

Finally, the fused template $t_g$ is generated by the mixed template and origin template combined with $\alpha_2$ and $\beta_2$. The feature fusion can be expressed as 
\begin{align}\label{ttt}
t_g=\alpha_2*t_m+\beta_2*t_o.
\end{align}
\subsection{Target-aware Contextual Attention Knowledge Distiller}
Knowledge distillation is a widely used model compression technique that helps the lightweight model balance between speed and performance.
Existing works \cite{yang2022focal,shen2021distilled,yang2024light} utilize attention maps from the teacher network to guide the student network in focusing on key regions. FGD \cite{yang2022focal} helps the student network in learning the most critical information by distilling channel and spatial attention maps. However, this distillation method is not well-suited for tracking tasks, as detection tasks require attention to focus on all targets in the image, while tracking is a target-specific task that only should focus on one target. For target tracking tasks, DST \cite{shen2021distilled} utilizes the attention weight map generated from the correlation operation between the template frame and the search frame, applying it to the feature map to suppress background information in the search frame and make the student mimic the response map. However, the similarity response map obtained from the correlation operation between the template frame and the search frame of the teacher network only matches from a local perspective, leading the student network to learn information that is influenced by similar distractors.

To accurately transfer teacher knowledge to student, we propose TCAKD, which distills attention map knowledge from a global perspective, reducing the impact of distracting features on the learning process of the student network. Instead of relying on local correlation operations to generate an attention map, TCAKD employs a Transformer-based structure to model the long-range dependencies between the template and the search frame, resisting distractor interference and enhancing target salience. The features are aggregated into a response map through summation, guiding the student to mimic this map based on the assumption that higher activation means greater importance. 

\begin{figure}[!t]
\centering
\setlength{\abovecaptionskip}{-0.1em}
\includegraphics[width=3.5in,keepaspectratio]{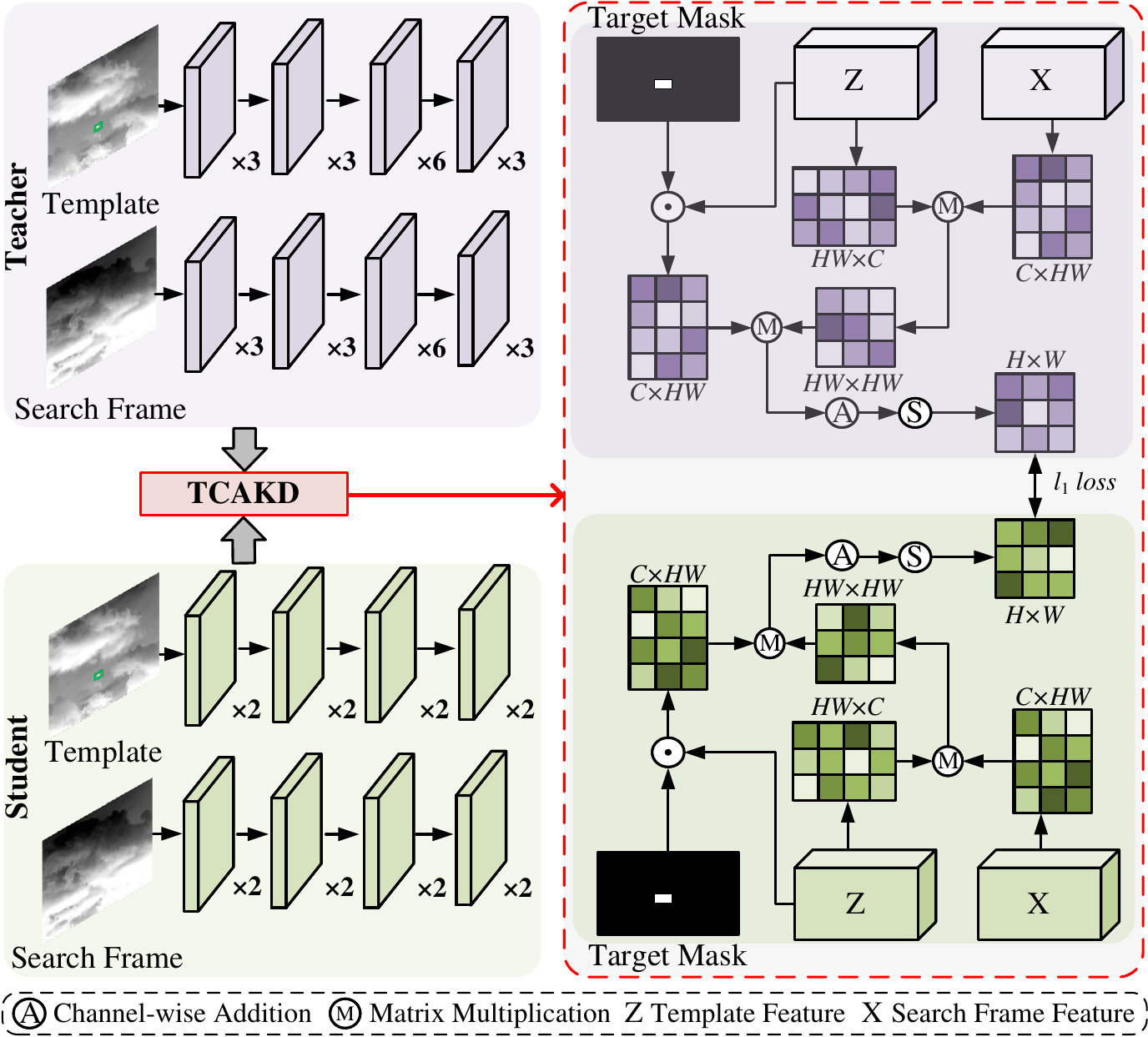}
\caption{Overview of the proposed TCAKD. The teacher network and student network are wrapped in purple and green boxes, respectively. The output feature maps of both template and search frame from each matching stage will all be computed for attention-based distillation. The target mask is generated from template, where the black area signifies the background and the white area indicates the target. }
\label{tcakd}
\end{figure}
The constructed module is shown in Fig \ref{tcakd}. The output from the corresponding stages will be used for attention-based distillation. Let ${F_z^t}\in\mathbb{R}^{{C}\times{H}\times{W}}$ and ${F_x^t}\in\mathbb{R}^{{C}\times{H}\times{W}}$, represent the teacher template feature map and search frame feature map from the backbone network, respectively. ${F_m}\in\mathbb{R}^{{H}\times{W}}$ denotes the target mask where the target region pixel value equals one and the background region pixel value equals zero. We demonstrate the calculation of attention map of teacher model as an example. We first multiply the target mask ${F_m}\in\mathbb{R}^{{H}\times{W}}$ with the template frame features ${F_z^t}\in\mathbb{R}^{{C}\times{H}\times{W}}$ channel by channel to focus on the target region feature and ignore the background information. Then, we transform the ${F_z^t}\in\mathbb{R}^{{C}\times{H}\times{W}}$ as ${K}\in\mathbb{R}^{{HW}\times{C}}$ and ${V}\in\mathbb{R}^{{C}\times{HW}}$. Then, we convert the search frame feature ${F_x^t}\in\mathbb{R}^{{C}\times{H}\times{W}}$ into ${Q}\in\mathbb{R}^{{C}\times{HW}}$. We get the similarity matrix between query and key by matrix multiplication. After, we calculate target-aware feature map by multiplying the similarity matrix with the value. Finally, to get teacher target-aware contextual attention map ${A_t}\in\mathbb{R}^{{H}\times{W}}$ we gather the feature map by summation and pass through a sigmoid function. The important region will have high value and irrelevant region will have low value. Similarly, we can also compute the attention map ${A_s}\in\mathbb{R}^{{H}\times{W}}$ of the student network. The formulation can be defined as
\begin{align}\label{tcakd_attention}
A_t=\sigma\left(\sum_{i=1}^{C}{V}\otimes \mathrm{Softmax}\left(\frac{K\otimes (Q\cdot F_m)}{\sqrt{d}}\right)\right),
\end{align}
where $\sigma$ and $\otimes$ denote sigmoid function and matrix multiplication, respectively. 

After get ${A_t}\in\mathbb{R}^{{H}\times{W}}$ and ${A_s}\in\mathbb{R}^{{H}\times{W}}$, the learning process can be defined as
\begin{align}\label{tcakd_loss}
\mathcal{L}_{TCAKD}=\mathcal{D}\left({A_t}\in\mathbb{R}^{{H}\times{W}},{A_s}\in\mathbb{R}^{{H}\times{W}}\right),
\end{align}
where $\mathcal{D}$ is defined as a $l_1$-norm loss.

\section{Experiments} \label{section4}
\subsection{Datasets}
We select ten infrared image sequences, each of which represents common challenges encountered in anti-UAV scenarios, such as heavy clouds and similar backgrounds. Seq. 1-Seq. 5 and Seq. 7 are collected from public datasets and can be accessed online \cite{2023CVPRAntiUAVDatasets,zhao20212nd}. The rest are collected by ourselves and carefully labeled. We use a total of 19392 images that cover targets of various scales. The division ratio of the training and testing sets is set to 7:3. 

\subsection{Evaluation Metrics and Experimental Setups}
The proposed network is trained by an SGD optimizer with a momentum of 0.9 and a weight decay of 0.0001. We use a cosine learning rate schedule, where warmup epochs are 50 and the base learning rate is set to 0.003. The total training epoch and batch are 250 and 12, respectively. We use random flipping for data augment. The network is trained from scratch without loading pre-training parameters. For training TCAKD, we use SiamCAP \cite{fang2024online} as teacher model and ResNet18 \cite{He_2016_CVPR} as a student model. The framework is implemented on a server with a NVIDIA GeForce RTX 3090 GPU accelerated by CUDA 11.1. For software implementation, we use Python 3.8 and Pytorch 1.8.1. The size of the search image is 384$\times$384. 

We select GlobalTrack \cite{2020AAAIHuang}, SiamYOLO \cite{2021ICCVWFang}, Unicorn \cite{2022ECCVYan} SimTrack \cite{chen2022backbone}, HCAT \cite{chen2022efficient}, ROMTrack \cite{cai2023robust}, HIT \cite{kang2023exploring}, SMAT \cite{gopal2024separable},  ODTrack \cite{zheng2024odtrack}, and MixFormer \cite{cuimixformer} for comparison with our method. SiamYOLO is designed for IRUT tracking task. SimTrack, HCAT, HIT, and SMAT belong to efficient tracking methods. GlobalTrack, SiamYOLO, and Unicorn are global search based tracking methods. ROMTrack, ODTrack, and MixFormer are high performance Transformer-based methods. All experiments are trained on the same infrared UAV datasets and tuned to the optimal.

We use the precision plot \cite{wu2013online}, success plot \cite{wu2013online}, normalized precision (NPrecision) plot \cite{muller2018trackingnet}, and state accuracy (SA) \cite{2021TMMJiang} to evaluate the performance of each method. We employ the precision at a 5 pixels threshold as a key metric for each tracker, which effectively assesses the tracking performance of small IRUTs \cite{kou2023infrared}. The normalized precision is derived by adjusting the precision relative to the dimensions of the ground truth bounding box. To rank each tracker in the normalized precision plot, we utilize the area under the curve (AUC) as a representative score. Additionally, the success plot computes the proportion of instances where the intersection over union (IoU) between the predicted and ground-truth bounding boxes exceeds a specified threshold. We apply the AUC as a ranking score for each tracker.

\subsection{Quantitative Results}




\begin{table*}[t!]
	\centering
	\caption{Quantitative Comparisons of the Proposed Method and Other Methods. The best three results are shown in \textcolor{red}{red}, \textcolor{green}{green}, and \textcolor{blue}{blue} fonts, respectively.}
	\vspace{-0.5em}
	\setlength\tabcolsep{2.5pt} 
	\fontsize{7.7}{8.5}\selectfont  
	\begin{tabular}{c|c|p{5.1em}<{\centering}|p{5.4em}<{\centering}|p{4.7em}<{\centering}|p{4.7em}<{\centering}|p{4.7em}<{\centering}|p{4.6em}<{\centering}|p{4.9em}<{\centering}|p{4.4em}<{\centering}|p{4.5em}<{\centering}|p{5.1em}<{\centering}|p{3em}<{\centering\arraybackslash}} \hline
\# Seq. & Metrics & {\makecell{GlobalTrack\\ (AAAI'20)}} & \makecell{SiamYOLO\\ (ICCVW'21)} & \makecell{Unicorn\\(ECCV'22)} & \makecell{HCAT\\(ECCV'22)} & \makecell{SimTrack\\(ECCV'22)} & \makecell{ROMTrack\\(ICCV'23)} & \makecell{SMAT\\(WACV'24)} & \makecell{HIT\\(ICCV'23)} & \makecell{ODTrack\\(AAAI'24)} & \makecell{MixFormer\\(TPAMI'24)}  & Ours \\  \hline
		\multirow{5}*{1} & SA & 0.661 & \textcolor{blue}{\textbf{0.700}} & \textcolor{green}{\textbf{0.716}} &0.235 & 0.215 & 0.379 & 0.213 & 0.216 & 0.584 & 0.416 & \textcolor{red}{\textbf{0.757}} \\
		& Success & 0.644 & \textcolor{blue}{\textbf{0.682}} &\textcolor{green}{\textbf{0.698}} &0.296 & 0.271 & 0.478 & 0.269 & 0.272 & 0.636 & 0.525 & \textcolor{red}{\textbf{0.790}} \\
		& Precision  & 0.869 & \textcolor{blue}{\textbf{0.886}} & \textcolor{green}{\textbf{0.909}}&0.285 & 0.280 & 0.488 & 0.277& 0.285 & 0.741& 0.533 & \textcolor{red}{\textbf{0.941}}\\
  & NPrecision  & 0.787 & \textcolor{blue}{\textbf{0.801}} & \textcolor{green}{\textbf{0.817}}&0.269 & 0.253 & 0.442 & 0.253& 0.255 & 0.681 & 0.485 & \textcolor{red}{\textbf{0.889}}\\
		& FPS & 9.2 & \textcolor{blue}{\textbf{60.3}} & 28.9&52.4 & \textcolor{red}{\textbf{63.1}} & 51.1 & 50.3 & \textcolor{green}{\textbf{62.1}} & 40.1 & 28.4 & 52.8\\	\hline
  
		\multirow{5}*{2} & SA & 0.415 & \textcolor{red}{\textbf{0.596}} &0.544 &0.046 & 0.037 & 0.350 & 0.017 & 0.044 & 0.037 & \textcolor{blue}{\textbf{0.580}} & \textcolor{green}{\textbf{0.590}} \\
		& Success & \textcolor{blue}{\textbf{0.402}} & \textcolor{red}{\textbf{0.580}} & 0.528&0.045 & 0.037 & 0.345 & 0.017 & 0.044 & 0.083 & \textcolor{green}{\textbf{0.572}} & \textcolor{red}{\textbf{0.580}} \\
		& Precision  & 0.621 & 0.789 & \textcolor{green}{\textbf{0.813}}&0.059 & 0.045 & 0.480 & 0.028 & 0.059 & 0.112 &\textcolor{green}{\textbf{0.795}}& \textcolor{red}{\textbf{0.827}}\\
  & NPrecision & 0.532 & \textcolor{green}{\textbf{0.709}} & 0.691&0.055 &0.048& 0.423 &0.029 & 0.054 & 0.104 & \textcolor{blue}{\textbf{0.696}} & \textcolor{red}{\textbf{0.740}}\\
		& FPS & 9.7 & \textcolor{blue}{\textbf{60.0}} & 29.1&52.6 & \textcolor{red}{\textbf{65.4}} & 52.3 & 55.4 & \textcolor{green}{\textbf{64.0}}& 44.2 & 28.7& 53.0\\ 
		\hline
		
		\multirow{5}*{3} & SA & 0.668 & \textcolor{green}{\textbf{0.691}} & \textcolor{blue}{\textbf{0.688}}&0.055 & 0.091 & 0.185 & 0.141 & 0.146 & 0.156 & 0.160 & \textcolor{red}{\textbf{0.728}} \\
		& Success & 0.652 & \textcolor{green}{\textbf{0.679}} & \textcolor{blue}{\textbf{0.673}}&0.068 & 0.112 & 0.230 & 0.173 & 0.179 & 0.191 &0.198 & \textcolor{red}{\textbf{0.751}}\\
		& Precision & \textcolor{blue}{\textbf{0.851}} & 0.835 & \textcolor{green}{\textbf{0.848}}&0.067 & 0.131 & 0.253 & 0.184 & 0.189 & 0.205 & 0.224& \textcolor{red}{\textbf{0.899}}\\
  & NPrecision & \textcolor{green}{\textbf{0.786}} & 0.769 & \textcolor{blue}{\textbf{0.773}}&0.067 & 0.114 & 0.225& 0.166 &0.169 & 0.185 & 0.195 & \textcolor{red}{\textbf{0.844}}\\
		& FPS & 9.6 & \textcolor{blue}{\textbf{61.4}} & 28.9&52.8 & \textcolor{red}{\textbf{69.2}} & 52.9 & 50.3 & \textcolor{green}{\textbf{61.9}} & 43.2 & 29.2 & 53.1\\ 
		\hline
		
		\multirow{5}*{4} & SA & 0.646 & \textcolor{green}{\textbf{0.665}} & \textcolor{blue}{\textbf{0.653}}&0.175 & 0.168 & 0.234 & 0.156 & 0.166& 0.158 & 0.173 &\textcolor{red}{\textbf{0.667}}\\
		& Success & 0.629 & \textcolor{green}{\textbf{0.662}} & \textcolor{blue}{\textbf{0.636}}&0.171 & 0.164 &0.229& 0.153 & 0.163 & 0.155 & 0.169 & \textcolor{red}{\textbf{0.652}}\\
		& Precision  & \textcolor{blue}{\textbf{0.885}} & 0.846 & \textcolor{green}{\textbf{0.906}}&0.209 & 0.214 &0.300 & 0.209 & 0.220 & 0.204 & 0.220 & \textcolor{red}{\textbf{0.856}}\\
  & NPrecision  & \textcolor{green}{\textbf{0.799}} & 0.772 & \textcolor{blue}{\textbf{0.796}}&0.199 & 0.193 &0.271 & 0.186 & 0.196 & 0.186 & 0.201 & \textcolor{red}{\textbf{0.810}}\\
		& FPS & 9.7 & \textcolor{blue}{\textbf{59.4}} & 28.9&52.7 & \textcolor{red}{\textbf{66.9 }}& 52.4 & 54.4 & \textcolor{green}{\textbf{63.5}} & 43.5 & 29.2 & 52.5\\ 
		\hline
		
		\multirow{5}*{5} & SA &0.357& \textcolor{blue}{\textbf{0.456}} & \textcolor{green}{\textbf{0.530}}&0.045 & 0.063 & 0.093 & 0.064 & 0.052 & 0.133 & 0.087 & \textcolor{red}{\textbf{0.597}}\\
		& Success & 0.347 & \textcolor{blue}{\textbf{0.443}} &\textcolor{green}{\textbf{ 0.514}}&0.044 & 0.060 & 0.089 & 0.0613 & 0.050 & 0.129 & 0.084 & \textcolor{red}{\textbf{0.579}}\\
		& Precision  & 0.531 & \textcolor{blue}{\textbf{0.667}} & \textcolor{green}{\textbf{0.811}}&0.067 & 0.107 & 0.144 & 0.104 & 0.077 &0.219 & 0.139 & \textcolor{red}{\textbf{0.856}}\\
  & NPrecision &0.455 & \textcolor{blue}{\textbf{0.563}} & \textcolor{green}{\textbf{0.680}}&0.061 & 0.086 & 0.121 & 0.087 & 0.068 & 0.181 & 0.117 & \textcolor{red}{\textbf{0.749}}\\
		& FPS & 9.8 &\textcolor{blue}{\textbf{ 59.6}} & 29.1&52.7 & \textcolor{red}{\textbf{69.5}} &52.4 & 52.8 & \textcolor{green}{\textbf{63.1}} & 42.7 & 28.9 & 50.4\\ 
		\hline
		
		\multirow{5}*{6} & SA & 0.733 & 0.731 & 0.733& \textcolor{red}{\textbf{0.756}} & 0.724 & 0.718 &0.734& 0.716 & 0.732 & \textcolor{blue}{\textbf{0.734}} & \textcolor{green}{\textbf{0.742}}\\
		& Success & 0.713& 0.711 & 0.713&\textcolor{red}{\textbf{0.736}} & 0.704 &0.699 & \textcolor{blue}{\textbf{0.714}} & 0.696 & 0.712 & 0.714 & \textcolor{green}{\textbf{0.723}}\\
		& Precision  & 0.987 & 0.979 & \textcolor{red}{\textbf{1.000}}&\textcolor{green}{\textbf{0.999}} & \textcolor{red}{\textbf{1.000}} & \textcolor{red}{\textbf{1.000}} & \textcolor{red}{\textbf{1.000}} & \textcolor{red}{\textbf{1.000}} &\textcolor{blue}{\textbf{ 0.997}} & \textcolor{red}{\textbf{1.000}} & \textcolor{red}{\textbf{1.000}}\\
   & NPrecision  &\textcolor{green}{\textbf{0.909 }}& 0.904 & \textcolor{blue}{\textbf{0.906}}&0.903 & 0.897 & 0.890&0.900 & 0.890 & 0.897 & 0.900 & \textcolor{red}{\textbf{0.910}}\\
		& FPS & 9.6 & \textcolor{blue}{\textbf{63.0}}& 29.0& 52.8 & \textcolor{red}{\textbf{65.1}} & 51.9& 53.5 & \textcolor{green}{\textbf{64.6 }}& 43.0 & 27.8 & 50.6\\ 
		\hline
		
		\multirow{5}*{7} & SA &0.856 & \textcolor{blue}{\textbf{0.866}} &\textcolor{green}{\textbf{ 0.872}}&0.266 & 0.379 & 0.612 & 0.265 & 0.702 & 0.789 & 0.515 & \textcolor{red}{\textbf{0.917}}\\
		& Success & 0.836 &\textcolor{blue}{\textbf{ 0.847}} & \textcolor{green}{\textbf{0.850}}&0.257 & 0.368 & 0.595 & 0.256 & 0.684 & 0.769 & 0.500& \textcolor{red}{\textbf{0.897}}\\
		& Precision  & \textcolor{green}{\textbf{0.996}} & 0.948 & \textcolor{blue}{\textbf{0.994}}&0.276 & 0.402 & 0.672& 0.292 & 0.752 &0.866 & 0.566 & \textcolor{red}{\textbf{0.992}}\\
  & NPrecision  & \textcolor{green}{\textbf{0.947}} & \textcolor{blue}{\textbf{0.914 }}& \textcolor{green}{\textbf{0.947}}&0.289 & 0.425 &0.682 & 0.292& 0.770& 0.868 & 0.573 & \textcolor{red}{\textbf{0.964}}\\
		& FPS & 9.6 & \textcolor{blue}{\textbf{60.6}} & 28.9&52.4 &\textcolor{red}{\textbf{ 67.5}} &51.7 & 47.6 & \textcolor{green}{\textbf{61.6}} & 42.3 & 27.5& 52.1\\ 
		\hline
		
		\multirow{5}*{8} & SA &\textcolor{green}{\textbf{0.816}} & 0.806 & \textcolor{blue}{\textbf{0.815}}&0.230 & 0.224 & 0.227 & 0.225 & 0.551 & 0.810 & 0.279 & \textcolor{red}{\textbf{0.818}}\\
		& Success & \textcolor{green}{\textbf{0.796}} & 0.787 &\textcolor{blue}{\textbf{0.795}}& 0.225 & 0.219 & 0.221 & 0.219 & 0.537 & 0.790 & 0.272 & \textcolor{red}{\textbf{0.798}}\\
		& Precision & \textcolor{green}{\textbf{0.998}} &0.964& \textcolor{red}{\textbf{1.000}}&0.268 &0.270& 0.272 & 0.270 & 0.686 & 0.983& 0.333 & \textcolor{blue}{\textbf{0.987}}\\
  & NPrecision  & \textcolor{green}{\textbf{0.943}} & 0.911& \textcolor{blue}{\textbf{0.931}}&0.258 & 0.253 &0.255 &0.253 & 0.635 & 0.924 & 0.315 & \textcolor{red}{\textbf{0.949}}\\
		& FPS & 9.5 & \textcolor{green}{\textbf{62.0}} & 28.8&52.6 & \textcolor{red}{\textbf{65.2}} & 52.5 & 54.4 & \textcolor{blue}{\textbf{60.8}} & 41.1 & 28.5& 52.7\\ 
		\hline

		\multirow{5}*{9} & SA & 0.523 & \textcolor{green}{\textbf{0.613}} & \textcolor{blue}{\textbf{0.612}}&0.090 & 0.103 & 0.114 & 0.070 & 0.128 & 0.104& 0.114 & \textcolor{red}{\textbf{0.792}}\\
		& Success &0.507 & \textcolor{green}{\textbf{0.597}} & \textcolor{blue}{\textbf{0.594}}&0.086 & 0.098 & 0.108 &0.067 &0.122 & 0.099 &0.108 & \textcolor{red}{\textbf{0.772}}\\
		& Precision & \textcolor{blue}{\textbf{0.800}} & 0.781 & \textcolor{green}{\textbf{0.917}}&0.197 & 0.225 & 0.242 & 0.156 &0.272 & 0.236& 0.242 & \textcolor{red}{\textbf{1.000}}\\
   & NPrecision  &0.650 & \textcolor{blue}{\textbf{0.720}} & \textcolor{green}{\textbf{0.773}}&0.178 & 0.162 & 0.182 & 0.122 & 0.207 & 0.170 & 0.183 & \textcolor{red}{\textbf{0.922}}\\
		& FPS & 9.5 & \textcolor{blue}{\textbf{63.0}} & 29.0&52.6 & \textcolor{red}{\textbf{69.8}} & 52.7& 51.5 & \textcolor{green}{\textbf{63.2}} & 43.1 & 27.6& 53.7\\ 
		\hline

		\multirow{5}*{10} & SA & 0.383 & 0.600 & \textcolor{green}{\textbf{0.653}}&0.411 & 0.602 & 0.403 & 0.396 & 0.601 & 0.596 & \textcolor{blue}{\textbf{0.606}} & \textcolor{red}{\textbf{0.746}}\\
		& Success & 0.373 & 0.583 & \textcolor{green}{\textbf{0.633}}&0.400 & \textcolor{blue}{\textbf{0.583}} &0.392 & 0.385 & 0.582 &0.578 &\textcolor{blue}{\textbf{0.587}} & \textcolor{red}{\textbf{0.727}}\\
		& Precision & 0.515 & 0.793 &0.962&0.574 & 0.988 & 0.574 & 0.574& \textcolor{blue}{\textbf{0.994}} & 0.979 &\textcolor{green}{\textbf{0.997}} & \textcolor{red}{\textbf{1.000}}\\
  & NPrecision  & 0.469 & 0.712 &0.813& 0.524 &0.803 & 0.502 &0.504& \textcolor{green}{\textbf{0.815}} &\textcolor{green}{\textbf{ 0.815}} &\textcolor{blue}{\textbf{ 0.814}} & \textcolor{red}{\textbf{0.905}}\\
		& FPS & 9.5 &\textcolor{green}{\textbf{ 61.1}} & 28.8&52.6 & \textcolor{blue}{\textbf{62.0}} & 50.2 & 50.7 & \textcolor{red}{\textbf{62.1}} & 43.2 & 28.4 & 52.8\\ 
		\hline
	\end{tabular}
	\label{table_different_methods}
\end{table*}

Table \ref{table_different_methods} presents a quantitative comparison between our method SiamDFF and other ten methods across ten infrared image sequences. It is evident that SiamDFF surpasses the competing methods in terms of SA, success, precision, and nprecision in most of the test sequences, especially for Seqs. 5, 9, and 10, where the target scale is relatively small, while maintaining a high inference speed of over 50 FPS. This demonstrates that our method can effectively track small targets in complex backgrounds, while other methods struggle to accurately capture these targets due to their inability to enhance infrared target features. In comparison to efficient tracking methods like SimTrack, HCAT, HIT, and SMAT, our SiamDFF significantly exceeds their performance across all metrics while still maintaining efficiency within an acceptable range. This can be attributed to the proposed TCAKD, which enhances the network's feature extraction capabilities without compromising inference speed. Specifically, compared with Unicorn which achieves the overall second-best quantitative results, our SiamDFF has improved mSA which generated from average the SA across all sequences by 5.3\%. 

\begin{figure*}[!thbp]
\centering
\setlength{\abovecaptionskip}{-0.1em}
\includegraphics[width=0.9\linewidth,keepaspectratio]{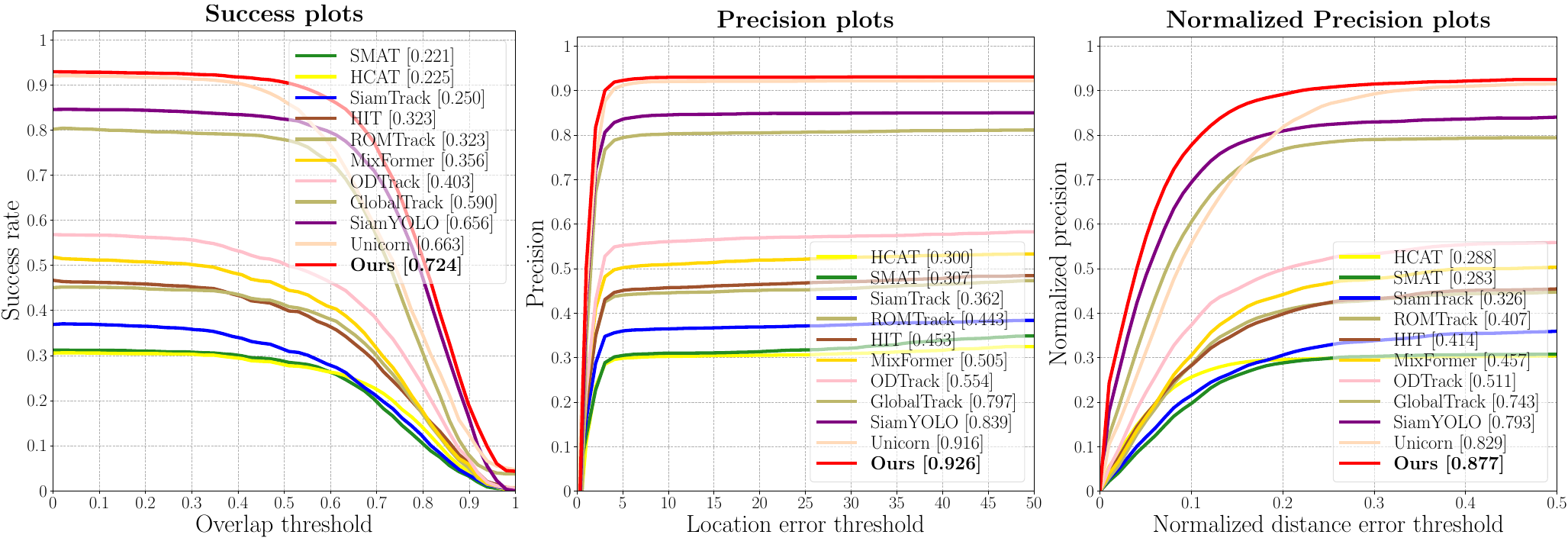}
\caption{Success plot, precision plot, and normalized precision plot of eleven trackers.}
\label{pr_suc_npr}
\end{figure*}

\begin{table*}[t!]
	\centering
	\caption{Computational complexity of different methods}
	\setlength\tabcolsep{2.5pt} 
	\fontsize{8.3}{10}\selectfont  
	\begin{tabular}{c|p{5em}<{\centering}|p{5em}<{\centering}|p{4em}<{\centering}|p{4em}<{\centering}|p{4em}<{\centering}|p{5em}<{\centering}|p{4em}<{\centering}|p{3.5em}<{\centering}|p{5em}<{\centering}|p{5em}<{\centering}|p{4em}<{\centering\arraybackslash}} \hline
    Metrics              & GlobalTrack  & SiamYOLO & Unicorn & HCAT  & SimTrack & ROMTrack  & SMAT & HIT & ODTrack & MixFormer & Ours \\ \hline
 \makecell{Params\\ (M)} & 58.76 & 7.12  & 121.42 & 6.27 & 0.28 & 25.40 & 2.59 & 37.58 & 92.12 & 35.97 & 6.82 \\  \hline
  \makecell{FLOPs\\ (G)} & 95.12 & 30.20 & 59.30  & 9.75 & 75.60& 92.10 & 1.56 & 14.18 & 73.10 & 33.18 & 58.48 \\  \hline
	\end{tabular}
	\label{computational_complexity}
\end{table*}

Furthermore, we present the success plot, precision plot, and normalized precision plot in Fig. \ref{pr_suc_npr}. A larger area under the curve means better tracking performance. Our method consistently demonstrates superior location accuracy and scale estimation compared to other methods. In particular, our method outperforms the second-best method Unicorn by 5.1\% in success, by 1\% in precision, and by 4.8\% in nprecision.

We also present the number of parameters and FLOPs for all comparative methods in the Table \ref{computational_complexity}. It can be observed that the proposed method has a moderate number of parameters and FLOPs compared to the other methods.

More experimental results can be found in the supplementary material.

\subsection{Qualitative Results}
\begin{figure*}[!htbp]
	\vspace{-0.3em}
	\centering
	\setlength{\abovecaptionskip}{-0.1em}
	\includegraphics[width=6.5in,keepaspectratio]{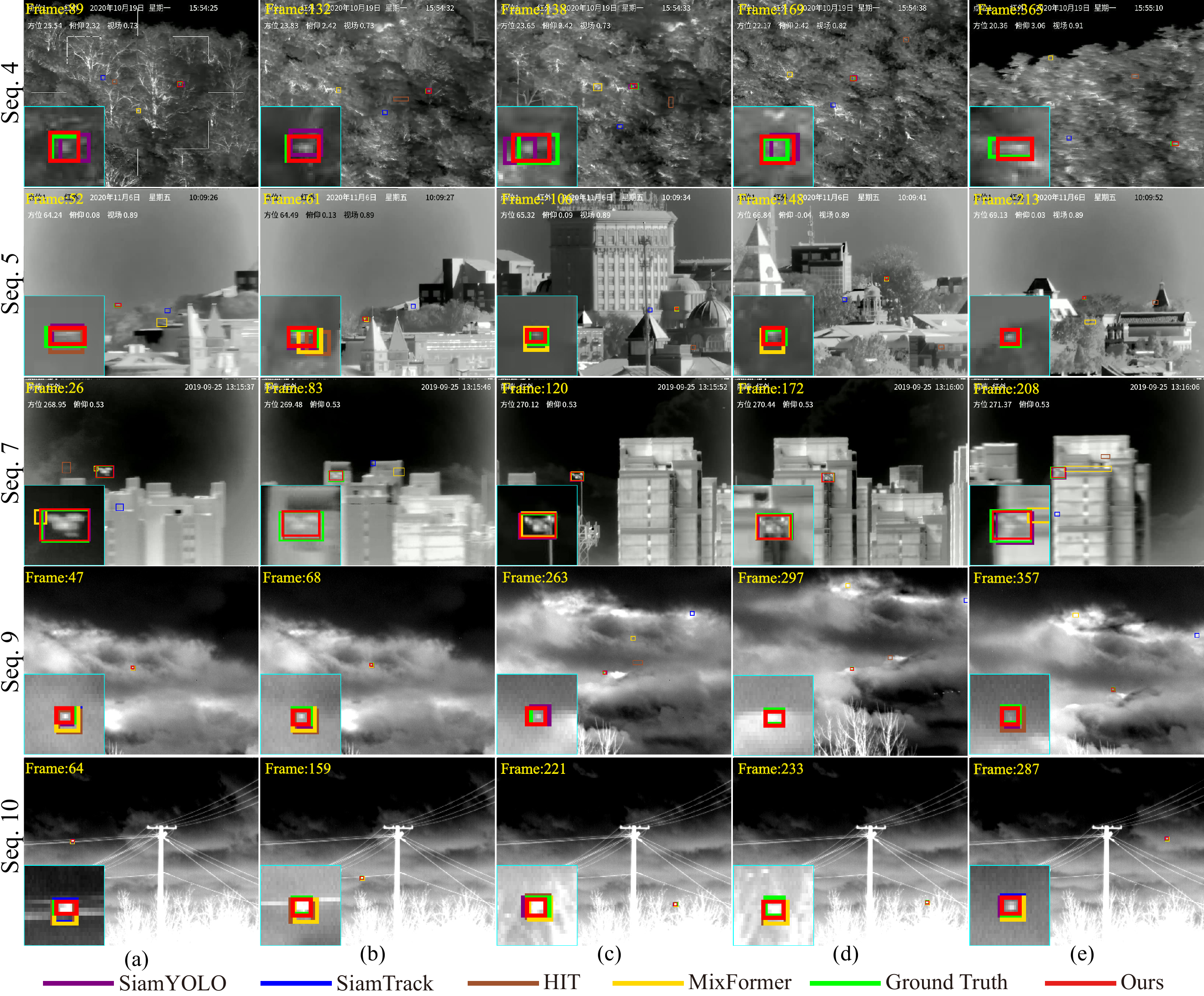}
	\caption{Qualitative tracking results of five methods from Seq. 4, Seq. 5, Seq. 7, Seq. 9, and Seq. 10.  Each row presents a sequence with five representative images. Close-ups are given for better visualization.}
	\label{tracking_results}
	\vspace{-0.5em}
\end{figure*}

Figure \ref{tracking_results} shows the infrared target tracking results of different methods from Seq. 4, Seq. 5, Seq. 7 Seq. 9, and Seq. 10. In Seq. 4, the UAV is small in scale and moves through a complex forest background. Local tracking methods can hardly track the target once a tracking failure occurs since UAV targets move fast and they lack a global search mechanism that limits the search range. SiamYOLO can only track the part of the target and mis-tracks the target during motion blur. In contrast, our method can consistently track the target. This is attributed to the DSFAM, which aggregates both global and local features to enhance the representational capability of the model. Additionally, the spatial attention mechanism within DSFAM effectively suppresses background features while enhancing the target features, thereby improving tracking performance in complex backgrounds. In Seq. 5, the UAV exhibits small and weak features, making it challenging for HIT and MixFormer to fully locate the target due to their limited feature extraction capabilities. In contrast, our model benefits from TCAKD, which enables the student model to effectively mimic the feature extraction abilities of the teacher model and allows our model to extract target features with greater precision. In Seq. 7(c) and (d), the UAV is large in size with a distinct shape. Although all methods can identify the UAV target correctly, our approach achieves superior localization accuracy. In Seq 7(a), (b), and (e), where the target appears blurred, our method still effectively estimates its scale. In Seq. 9(c), the target is completely obscured by bright clouds, and only our method successfully tracks the target, demonstrating the effectiveness of our approach in extracting IRUT features. In Seq. 10(c) and (d), as the UAV enters a bush, the target is blended with the background. Although HIT and MixFormer manage to locate the target to some degree, they struggle to estimate its size accurately. By employing both the original and mixed templates, DCFAM enhances the robustness of the model against distractors during the feature interaction process.
\begin{table}[!t]
\centering
 \vspace{-0.3em}
\caption{Effectiveness of integrating STEN, DSFAM, DCFAM, and TCAKD}
\fontsize{6.5}{10}\selectfont  
	\setlength\tabcolsep{0.4em}
 \begin{tabular}{c|c|c|c|c|c|c|c|c}\hline
\multirow{2}{*}{\makecell[c]{Row}}&\multirow{2}*{\makecell[c]{STEN}} & \multirow{2}{*}{\makecell[c]{DSFAM}} & \multirow{2}{*}{\makecell[c]{DCFAM}} & \multirow{2}{*}{\makecell[c]{TCAKD}} & \multicolumn{4}{c}{Metrics} \\ \cline{6-9}
		 & & & & & mSA & Success & Precision & NPrecision  \\  \hline
		1&- & - & - & - & 0.674 & 0.670 & 0.862 & 0.817\\
		2&\checkmark & - & - & - & 0.686 & 0.682 & 0.877 & 0.832\\
		3&- & \checkmark & - & - & 0.692 & 0.686 & 0.882 & 0.836\\
		4&- & - & \checkmark & - & 0.687 & 0.683 & 0.875 & 0.831\\
        5&- & - & - & \checkmark & 0.707 & 0.702 & 0.906 & 0.858\\
		6&\checkmark & \checkmark & - & - & 0.700 & 0.695 & 0.889 & 0.843\\
        7&\checkmark & \checkmark & \checkmark & - & 0.709 & 0.702 & 0.897 & 0.851\\
        \hline
        8&\checkmark & \checkmark & \checkmark & \checkmark & {\bf0.735} & {\bf0.724} & {\bf0.926} & {\bf0.877} \\    \hline
	\end{tabular}
	\label{table:ablation}
\end{table}
\subsection{Ablation Study}\label{ablation}

\subsubsection{Effectiveness of Integrating All Sub-Component}
Table \ref{table:ablation} presents the quantitative results of our tracking method with and without the proposed components: STEN, DFFAM, DCFAM, and TCAKD. The results in rows 1 and 8 demonstrate that compared to the baseline, mSA, success, precision, and nprecision have increased by 6.1\%, 5.4\%, 6.4\%, and 6\%, respectively. This clearly indicates that incorporating any of the proposed components enhances tracking performance relative to the baseline, highlighting that each component contributes to improving the capability of the model to track infrared UAV targets in complex backgrounds. Furthermore, the integration of all components yields superior results compared to single or paired combinations, suggesting that each component provides complementary advantages.

\subsubsection{Effectiveness of Integrating STEN}
\begin{table}[!t]
 \fontsize{8.7}{12}\selectfont  
 \centering
 \vspace{-0.3em}
 \caption{Ablation study of different thresholds}
\begin{tabular}{c c c c c c c c}\hline
  \makecell*[c] {Threshold} & mSA &  Success & Precision & NPrecision  \\ 
  \hline
  0.5 & 0.676 & 0.673 & 0.866 & 0821 \\ 
  \hline
  0.6 & 0.684 & 0.681 & 0.875 & 0.830 \\ 
  \hline
  0.7 & {\bf0.686} & {\bf0.682} & {\bf0.877} & {\bf0.832}\\ 
  \hline
   0.8 & 0.679 & 0.675 & 0.869 & 0.824 \\ 
  \hline
   1.0 & 0.674 & 0.670 & 0.862 & 0.817 \\ 
  \hline
\end{tabular}
 \label{sten_thres}   
\end{table}
As shown in rows 1 and 2 of Table \ref{table:ablation}, changing cross-attention to STEN brings an improvement of mSA by 1.2\%, success by 1.2\%, precision by 1.5\%, and nprecision by 1.5\%. This indicates that minimizing irrelevant context effectively enhances the features of the target regions. Unlike cross-attention which computes similarities across all query-key pairs, our proposed STEN focuses solely on the most relevant regions, thereby reducing background interference.

The key parameter for STEN is the specified threshold $T$, which significantly affects the considered part of background. Table \ref{sten_thres} comprehensively shows the performance change brought by $T$. When $T$ =1, STEN becomes native cross-attention. We find that STEN consistently outperforms cross-attention and achieves the best results when the threshold $T$ is carefully and manually selected as 0.7. We argue that when $T$ is small, only a limited amount of contextual information is utilized, which fails to effectively highlight the features of the target regions. Conversely, when $T$ is large, an excess of background information can affect the learning of the cross-attention values. We also attempt to set the threshold $T$ as a learnable parameter and found that the learned value, $T = 0.6426$, yields tracking performance very close to that of the manually selected optimal threshold $T = 0.7$. Further experimental details can be found in the supplementary material.

\subsubsection{Effectiveness of Integrating DSFAM}

\begin{table}[!t]
 \fontsize{8.7}{12}\selectfont  
 \centering
  \vspace{-0.3em}
 \caption{Ablation study for different fusion strategies of DSFAM}
\begin{tabular}{c c c c c c c c}\hline
  \makecell*[c] {Fusion Strategy} & mSA &  Success & Precision & NPrecision  \\ 
  \hline
  Concatenate &  0.679 & 0.676 & 0.865 & 0.820 \\ 
  \hline
  Sum  & 0.682 & 0.678 & 0.869 & 0.823 \\ 
  \hline
  Ours & {\bf0.692} & {\bf0.686} & {\bf0.882} & {\bf0.836} \\ 
  \hline
\end{tabular}
 \label{DSFAM-fusion-way}   
\end{table}

As shown in rows 1 and 3 of Table \ref{table:ablation}, the mSA, success, precision, and nprecision improve by 1.8\%, 1.6\%, 2\%, and 1.9\%, respectively. The Transformer-based branch focuses on extracting global features that capture contour details. In contrast, DSFAM integrates local features, which enrich the representation of the IRUT in the search frame. The combination of local and global features significantly enhances tracking performance. 

Table \ref{DSFAM-fusion-way} provides a comprehensive evaluation of the performance achieved through various fusion methods. Both direct addition and concatenation can enhance performance compared to baseline. This indicates that simultaneously integrating local and global features provides a more effective representation of target characteristics in complex scenarios. When employing attention-guided fusion, DSFAM achieves the highest performance. The ability to dynamically adjust the weights of different branches in response to various input images is emphasized, recognizing that different feature types contribute uniquely to targets of varying sizes.

\begin{figure}[!t]
	\centering
	\setlength{\abovecaptionskip}{-0.1em}
	\includegraphics[width=3.3in,keepaspectratio]{dsfamvisual.pdf}
	\caption{Visualization of response maps from three representative sequences before and after applying DSFAM. The features of the target region are significantly enhanced after leveraging DSFAM.}
	\label{dsfamvisual}
\end{figure}

Figure \ref{dsfamvisual} visualizes the feature maps before and after applying DSFAM from Seq. 6, Seq. 7, and Seq. 10. It is evident that before adding DSFAM, the network fails to focus on the target region. However, after incorporating local features, the target area is significantly enhanced. Additionally, the attention mechanism effectively mitigates background interference, leading to more salient target features. By leveraging this dual branch, the network is better equipped to accurately classify and locate the target in subsequent processing. 
\subsubsection{Effectiveness of Integrating DCFAM}
\begin{table}[!t]
 \fontsize{8.7}{12}\selectfont  
 \centering
  \vspace{-0.3em}
 \caption{Ablation study for different fusion strategies of DCFAM}
\begin{tabular}{c c c c c c c c}\hline
  \makecell*[c] {Fusion Strategy} & mSA &  Success & Precision & NPrecision  \\ 
  \hline
  Concatenate & 0.671 & 0.667 & 0.854 & 0.809 \\ 
  \hline
  Sum  & 0.677 & 0.673 & 0.865 & 0.821 \\ 
  \hline
  Ours & {\bf0.687} & {\bf0.683} & {\bf0.875} & {\bf0.831} \\ 
  \hline
\end{tabular}
 \label{DCFAM-fusion-way}   
\end{table}
From rows 1 and 4 of Table \ref{table:ablation}, we can observe that the incorporation of DCFAM results in an enhancement of 1.3\% in mSA, 1.3\% in success, 1.3\% in precision, and 1.4\% in nprecision. This indicates that relying solely on a mixed template of contextual information is insufficient for addressing the challenges in complex backgrounds. By utilizing the original template and the mixed template, DCFAM effectively minimizes the influence of excessive background information. Additionally, it allows the model to leverage the global contextual information from the search frame, enhancing its ability to adapt to new target shapes in the template frame. 

Table \ref{DCFAM-fusion-way} presents a thorough evaluation of performance across various fusion methods. It is clear that simple concatenation can actually diminish model performance, emphasizing the importance of a well-designed fusion strategy to enhance overall effectiveness. Meanwhile, utilizing an addition way yields only a slight improvement in performance. In contrast, the use of DCFAM leads to a significant performance improvement. This demonstrates that employing channel attention based on local channel interactions effectively guides the fusion process, enriching the template information. 

\subsubsection{Effectiveness of Integrating TCAKD}
\begin{figure}[!t]
	\centering
	\setlength{\abovecaptionskip}{-0.1em}
	\includegraphics[width=3.3in,keepaspectratio]{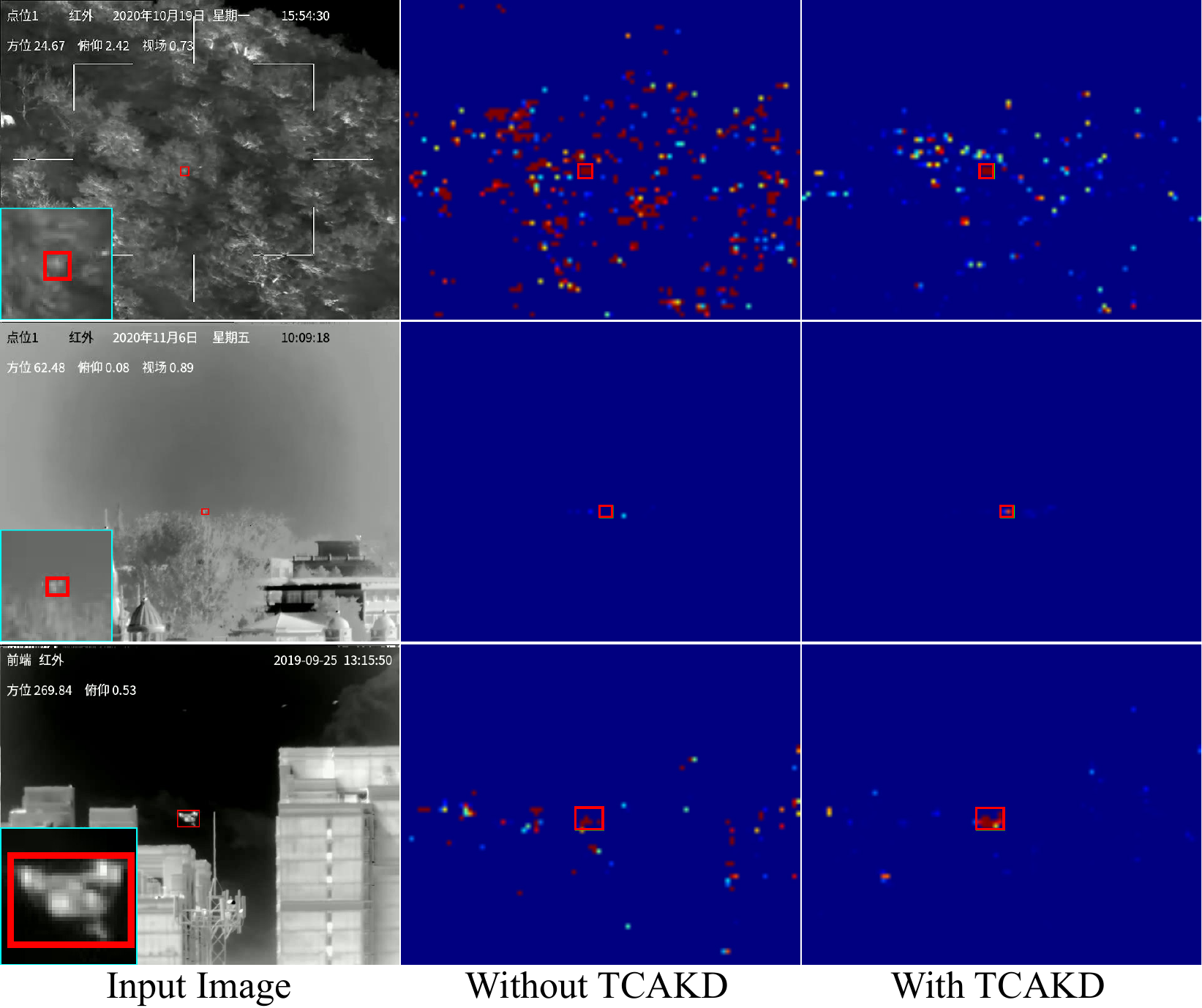}
	\caption{Visualization of target prior from three representative sequences before and after applying TCAKD. The red box indicates the tracking target. The network can better focus on the target region after leveraging TCAKD.}
	\label{tcakdvisual}
\end{figure}
From rows 1 and 8 of Table \ref{table:ablation}, we can observe that the incorporation of DCFAM results in an enhancement of 3.3\% in mSA, 3.2\% in success, 4.4\% in precision, and 4.1\% in nprecision, indicating TCAKD significantly enhances the backbone feature extraction capability by effectively transferring knowledge from the teacher network to the student network.

Figure \ref{tcakdvisual} visualizes the target prior \cite{2022ECCVYan} calculated between template and search frame in backbone network from Seq. 4, Seq. 5, and Seq. 7. In Seq. 4, the target is completely obscured by the background. Without TCAKD, the template matches with many similar objects, preventing the network from focusing on the target area. However, with TCAKD, the network can accurately concentrate on the target while significantly suppressing the background. In Seq. 5, the target is small and has weak features. The absence of TCAKD results in the network's failure to identify the target, whereas its inclusion allows for accurate tracking. Sequence 7 presents a large-scale target scenario. Without TCAKD, the network can only partially locate the target. However, with TCAKD, the network fully focuses on it. Overall, these observations highlight that TCAKD can be applied across various scales while maintaining strong focus, even in complex backgrounds.

\section{Further Discussion and Analysis} \label{section5}
\subsection{Comparative Analysis of Our SiamDFF and Conventional Siamese Architectures}
Traditional Siamese networks rely on global similarity computation between the template and search frame, making them highly vulnerable to background interference when tracking low-resolution and weak infrared targets. To address this, our proposed SiamDFF architecture introduces three dedicated modules. STEN employs intensity-aware attention to retain only the top-k of salient query-key pairs, effectively suppressing irrelevant context. DSFAM combines global Transformer features and local CNN details via spatial attention, enhancing multi-scale target representation. DCFAM adaptively fuses original and context-enhanced templates using channel-wise attention without dimensional compression, preserving semantic integrity. Overall, this design enhances feature discrimination, improves robustness under low-contrast conditions, and increases adaptability to scale variations, offering a principled solution for challenging infrared UAV tracking scenarios.

\subsection{Analysis of Intensity-aware Multi-head Cross-attention (IMC) Module}
The proposed Intensity-aware Multi-head Cross Attention (IMC) module enhances low-contrast infrared targets by amplifying salient regions and suppressing irrelevant context. Unlike standard cross-attention, IMC applies a top-k thresholding strategy to retain only the most informative query-key pairs, producing a sparse and focused attention map. This not only strengthens weak yet meaningful features but also filters out distractive background noise—particularly vital for thermal imagery with weak textures. Ablation results (Tables \ref{computational_complexity} and \ref{table:ablation}) confirm that IMC consistently outperforms standard cross-attention, validating its effectiveness in boosting target saliency and tracking performance.

\subsection{Failure Cases and Limitations of Our Tracking Method}
\begin{figure}[!t]
\centering
\includegraphics[width=0.96\linewidth, keepaspectratio]{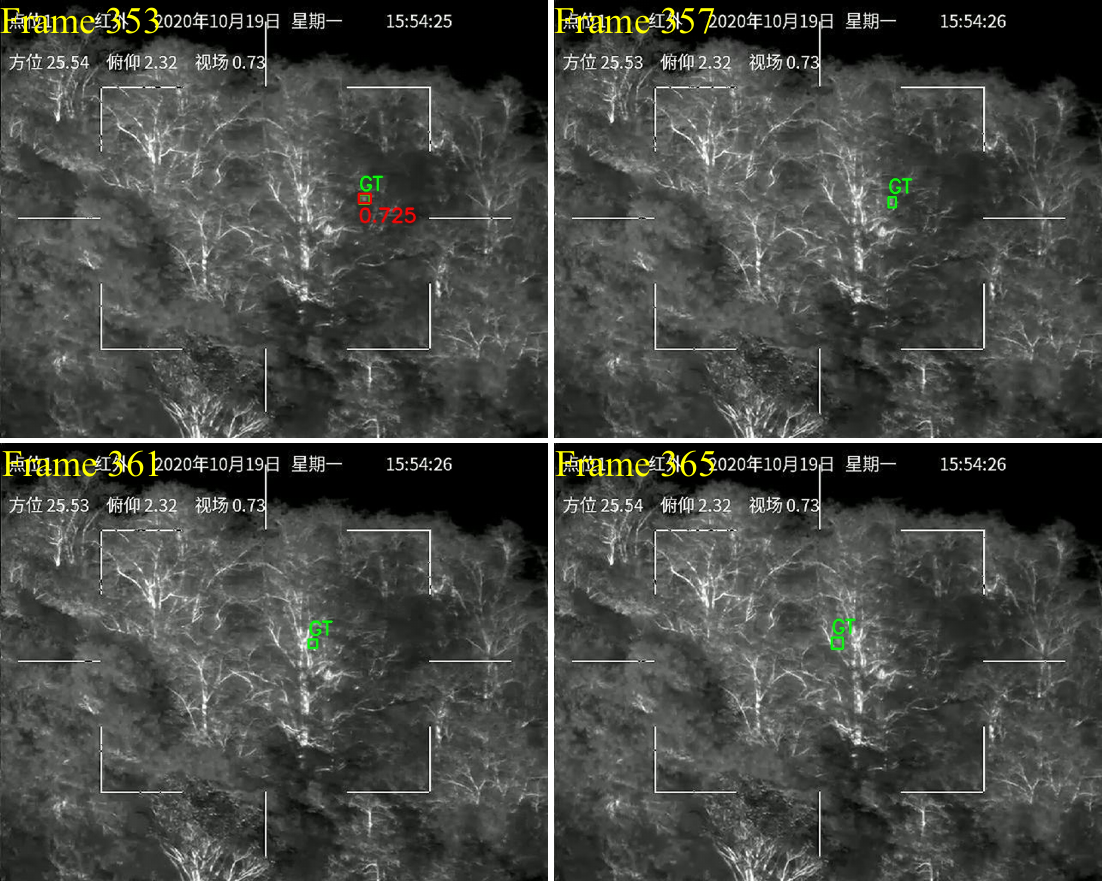}
\caption{Failure tracking results of our method on Seq. 4. Ground truth is shown in green, and tracking results are shown in red.}
\label{failure_case_track}
\end{figure}
As presented in Fig. \ref{failure_case_track}, the UAV target features are extremely weak, our tracking method fails to track the targets correctly in frames 357, 361, and 365 of Seq. 4. This means that the proposed method may encounter tracking failure when the target features are extremely weak, the target undergoes deformation, or historical templates are not used. This highlights a limitation that the proposed method should address in future work.

\section{Conclusion} \label{section6}
In this paper, we introduce a dynamic feature fusion Siamese network (called SiamDFF) for real-time IRUT tracking in complex backgrounds. First, we present a novel module STEN designed to minimize the influence of excessive background information on cross-attention values. Second, we propose a module DSFAM that effectively combines global and local features using a spatial attention mechanism, enhancing the feature representation capability in the search frame. Third, we introduce a module DCFAM that integrates the original template while utilizing a mixed template to improve the contextual awareness of the target in the search frame, aiming to reduce the impact of distractions on the information delivered by the template. To further enhance tracking performance, we propose a knowledge distiller TCAKD to strengthen the feature extraction ability of the backbone by transferring the global contextual attention map produced by the teacher model. Extensive experiments demonstrate the effectiveness of our approach, achieving high tracking performance and enabling real-time tracking.

\section*{Acknowledgments}
The authors would like to thank Chenhui Zhang for his assistance with the experiments, and the reviewers for their valuable and insightful suggestions.


%

\ifCLASSOPTIONcaptionsoff
  \newpage
\fi

\bibliographystyle{IEEEtran}
\bibliography{reference}

\clearpage
\def\INSIDEMAIN{} 

\providecommand{\suppTitle}{\bf \LARGE Infrared UAV Target Tracking with Dynamic Feature Refinement and Global Contextual Attention Knowledge Distillation\\ \noindent{\bf{\fontsize{20}{25}\selectfont\textbf{------}}}Supplementary Material\noindent\bf{\fontsize{20}{25}\selectfont\textbf{------}}}

\ifdefined\INSIDEMAIN
  \twocolumn[
  \centering\fontsize{25}{10}\selectfont\section*{\suppTitle} 
   \vspace{1em}
]
  
\else
  \title{\suppTitle}
  \maketitle
\fi

\setcounter{section}{0}  
\setcounter{subsection}{0}  
\setcounter{subsubsection}{0}  

\setcounter{figure}{0}   
\setcounter{table}{0}    


In this part, we provide the supplementary material of our work that is not included in the article due to space limitations. We first introduce the details of the training and testing datasets that we use to train and test our network.

\section{Dataset Description}
In this section, we present the specific details of the dataset we used. The summary of the dataset, which is composed of ten image sequences representing different challenges in the task of IRUT tracking, is shown in Table \ref{dataset_table}. For Seq. 1-Seq. 5 and Seq. 7, they are infrared sequences from 2023 IEEE Conference on Computer Vision and Pattern Recognition 3rd anti-UAV Workshop \& Challenge \cite{2023CVPRAntiUAVDatasets}; For the remaining sequences, we use DJI Phantom 4 Pro V2.0 as the UAV target we aim to track. The frame size for each sequence is 640$\times$512.

\begin{figure*}[!thbp]
\centering
\includegraphics[width=\linewidth, keepaspectratio]{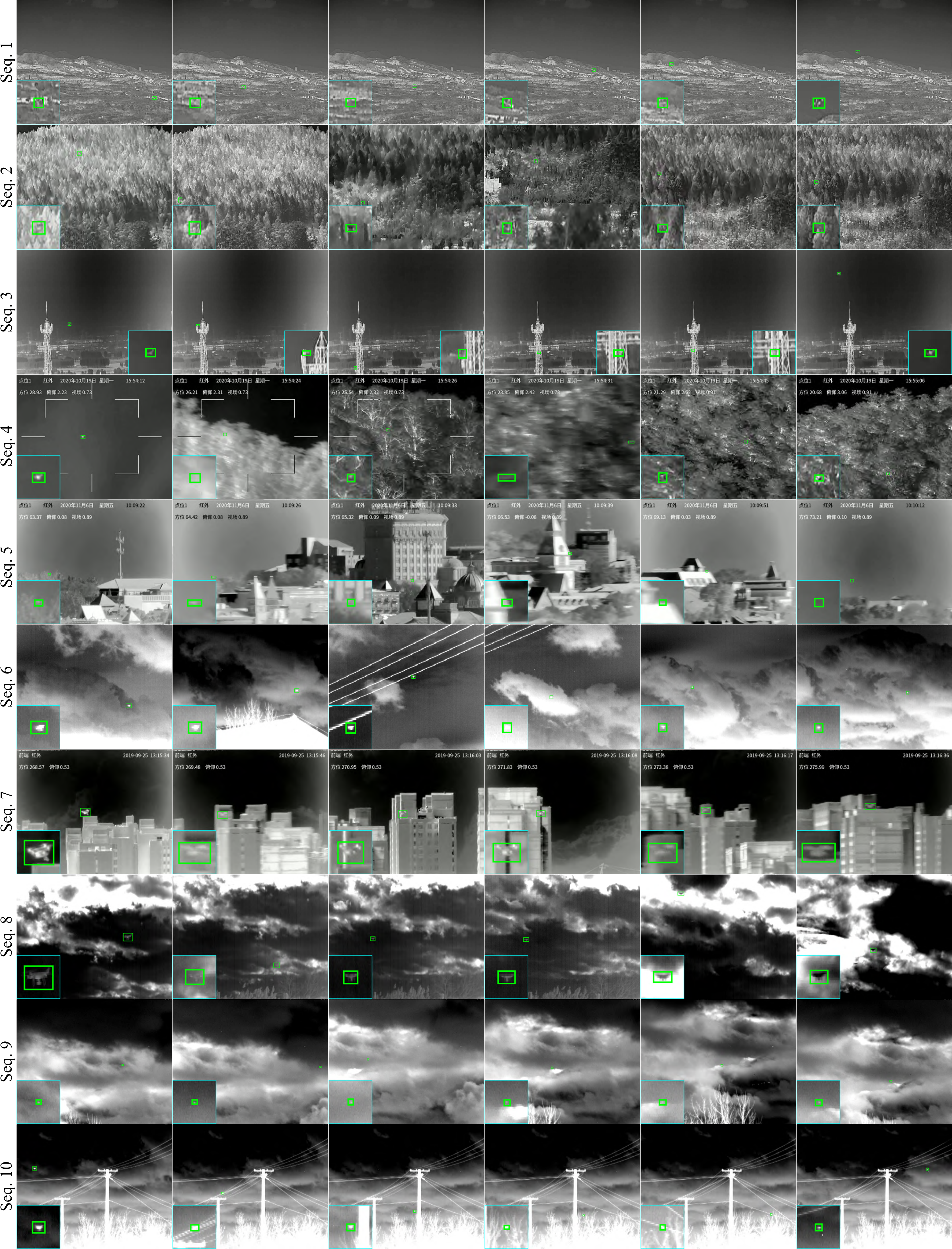}
\caption{Representative images of ten sequences selected as the dataset for training and testing. Each row presents six images from each sequence, respectively. The targets are bounded using green boxes, and close-ups are given in the left-bottom or right-bottom corner of each image.}
\label{representative_images}
\end{figure*}

\begin{table*}[!thbp]
	\centering
	\caption{Details of ten real test sequences}
		\fontsize{9.5}{9}\selectfont  
	\renewcommand{\arraystretch}{1.5}
	\begin{tabular}{c|c|c|c|c|c|c}		\hline
		\multirow{2}*{\# Seq.} & \multicolumn{3}{c|}{Number of frames} & \multirow{2}*{ \makecell[c]{Target size\\ (pixels)}} & \multirow{2}*{Target details} & \multirow{2}*{Background details}\\ 	\cline{2-4}
		& Train & Test & Total & & & \\ 	\hline
		1 & 869 & 375 & 1244  & 80-324 & Small size  & Mountains \\ \hline
		
    		2 & 1109 & 375 & 1484  & 36-594 & Small size& Complex tree background\\	\hline
		
		3 & 896 & 375 & 1271  & 15-272 & Small size & Sky and signal tower \\ \hline
		
		4 & 1118 & 373& 1491  & 70-1892 & Multi-scale variation and motion blur & Complex tree background \\ \hline
		
		5 & 1125 & 375 & 1500  & 28-220 & Small size and dim & Building background \\	\hline
		
		6 & 1275 & 1276 & 2551  & 28-567 & Small size & Strong cloud and electrical wires \\  \hline
		
		7 & 1500 & 500 & 2000  & 682-1426 & Motion blur and blend with background  & Building background\\ \hline
		
		8 & 1047 & 522 & 1569  & 120-1519 & Multi-scale variation & Sky and strong cloud\\  \hline
  
  9 & 2880 & 360 & 3240  & 20-110 & Small size and dim& Strong cloud\\  \hline
  
  10 & 2703 & 339 & 3042  & 6-304 & Small size & Bush background and electrical poles\\  \hline

	\end{tabular}
	\label{dataset_table}
\end{table*}

We also display six representative images of each sequence in Fig. \ref{representative_images}. For better visualization, we highlight the target within each image using a green bounding box and a close-up for each image. We briefly summarize the characteristics of ten sequences as follows. In Seq. 1, UAV moves near and far in the mountain and out of view frequently. In Seq. 2 and Seq. 4, the target is relatively small in size and has low contrast compared to the background. In Seq. 3, the UAV moves around a tower, which makes it prone to occlusion. In Seq. 5, the target is dim, small, and moves fast, leading to rapid changes in the background. In Seq. 6, the target is slightly brighter than the background and susceptible to strong clouds and electrical wires during motion. In Seq. 7, the target has a relatively large scale and moves between buildings while always blending into the building background, which can make it challenging to distinguish the target from the building background. In Seq. 8, the target undergoes multi-scale changes in size as it flies from near to far distances. In Seq. 9, the target scale is extremely small and appears as a bright spot. When flying in the clouds, the target is usually brighter than the background, but when flying into bright clouds, it can become difficult to distinguish the target from the background. In Seq. 10, the target object is relatively small, and as it flies towards the bush, the targets are immersed in the background.

\section{More Quantitative Results on infrared anti-UAV tracking dataset}

\subsection{Performance Comparison of Attribute-Based Trackers}
\begin{table}[!t]
\centering
\caption{Performance Comparison of attribute-based trackers in terms of mSA}
\fontsize{8.5}{10}\selectfont  
\setlength\tabcolsep{0.5em} 
 \begin{tabular}{c|c|c|c|c|c|c|c}\hline
\multirow{2}{*}{\makecell[c]{Method}}& \multicolumn{7}{c}{ Attributes} \\ \cline{2-8}
		 & LR & OC & FM & SV &TC&LI&OV \\  \hline
		GlobalTrack&0.334 & 0.437 & 0.687 & 0.697 &  0.580&0.697& 0.819\\
		SiamYOLO &0.485 &  0.586& 0.715 &0.691 & 0.651&0.727&0.838\\
		Unicorn&  0.552 &0.606 & 0.731 & 0.728 & 0.634&0.733&0.895\\
		HCAT &0.104&0.233 & 0.403& 0.447 & 0.140 &0.173&0.124\\
        SimTrack&0.189&0.329 &0.432 & 0.456&0.164 &0.175&0.117\\
		ROMTrack &0.137 &0.248  &0.468 &0.429&0.342 &0.180&0.280\\
        SMAT&0.100&0.220&0.398&0.435 & 0.132&0.162&0.126\\
        HIT&0.196 &0.344& 0.558 & 0.596& 0.262&0.379&0.139 \\   
        ODTrack&0.208 & 0.330&0.627&0.715& 0.344&0.522&0.239\\
         MixFormer&0.223 &0.349&0.481& 0.482 &0.358 &0.211&0.324\\
          SiamDFF&\bf{0.665} &\bf{0.721}& \bf{0.757} &\bf{0.737}& \bf{0.710}& \bf{0.807}& \bf{0.956}\\
        \hline
	\end{tabular}
	\label{msa}
\end{table}

\begin{table}[!t]
\centering
\caption{Performance Comparison of attribute-based trackers in terms of Success}
\fontsize{8.5}{10}\selectfont  
	\vspace{-0.1em}
\setlength\tabcolsep{0.5em} 
 \begin{tabular}{c|c|c|c|c|c|c|c}\hline
\multirow{2}{*}{\makecell[c]{Method}}& \multicolumn{7}{c}{Attributes} \\ \cline{2-8}
		 & LR & OC & FM & SV &TC&LI&OV \\  \hline
		GlobalTrack&0.227 & 0.347 & 0.540 & 0.522 &  0.469&0.651& 0.582\\
		SiamYOLO &0.354 &  0.482& 0.547 &0.528 & 0.543&0.692&0.648\\
		Unicorn&  0.419 &0.474 & 0.622 & 0.558 & 0.527&0.694&0.634\\
		HCAT &0.137&0.173 & 0.204& 0.244 & 0.110 &0.156&0.124\\
        SimTrack&0.211&0.252 &0.225 & 0.265&0.131 &0.158&0.117\\
		ROMTrack &0.188 &0.217  &0.276 &0.233&0.287 &0.165&0.280\\
        SMAT&0.141&0.181&0.205&0.235 & 0.123&0.143&0.126\\
        HIT&0.184 &0.281& 0.329 & 0.340& 0.199&0.330&0.139 \\   
        ODTrack&0.207 & 0.269&0.383&0.400& 0.274&0.444&0.239\\
         MixFormer&0.283 &0.327&0.304& 0.288 &0.309 &0.190&0.324\\
          SiamDFF&\bf{0.517} & \bf{0.565}&\bf{0.618}& \bf{0.604} &\bf{0.603}& \bf{0.780}& \bf{0.685}\\
        \hline
	\end{tabular}
	\label{success}
\end{table}

\begin{table}[!t]
\centering
\caption{Performance Comparison of attribute-based trackers in terms of Precision}
\fontsize{8.5}{10}\selectfont  
	\vspace{-0.1em}
\setlength\tabcolsep{0.5em} 
 \begin{tabular}{c|c|c|c|c|c|c|c}\hline
\multirow{2}{*}{\makecell[c]{Method}}& \multicolumn{7}{c}{Attributes} \\ \cline{2-8}
		 & LR & OC & FM & SV &TC&LI&OV \\  \hline
		GlobalTrack&0.502 & 0.622 & 0.885 & 0.910 &  0.772&0.906& 0.797\\
		SiamYOLO &0.653 &  0.762& 0.897 &0.909 & 0.794&0.888&0.825\\
		Unicorn&  0.851 &0.904 & 0.962 & 0.985 & 0.858&0.966&0.869\\
		HCAT &0.196&0.359  & 0.526& 0.599 & 0.176 &0.239&0.155\\
        SimTrack&0.383 &0.568 &0.596 & 0.661&0.210 &0.252&0.158\\
		ROMTrack &0.256 &0.392  &0.625 &0.601 &0.438 &0.260&0.380\\
        SMAT&0.189&0.347 &0.540& 0.600 & 0.172&0.223&0.170\\
        HIT&0.396 &0.598& 0.752 & 0.850& 0.322&0.517&0.188 \\   
        ODTrack&0.421 & 0.573 & 0.826& 0.976& 0.425&0.678&0.316\\
         MixFormer&0.439 &0.600&0.655& 0.690 &0.467 &0.296&0.443\\
          SiamDFF&\bf{0.922} & \bf{0.936}&\bf{0.952}& \bf{0.963} &\bf{0.878} & \bf{0.981}& \bf{0.871}\\
        \hline
	\end{tabular}
	\label{precision}
\end{table}

\begin{table}[!t]
\centering
\caption{Performance Comparison of attribute-based trackers in terms of NPrecision}
\fontsize{8.5}{10}\selectfont  
	\vspace{-0.1em}
\setlength\tabcolsep{0.5em} 
 \begin{tabular}{c|c|c|c|c|c|c|c}\hline
\multirow{2}{*}{\makecell[c]{Method}}& \multicolumn{7}{c}{Attributes} \\ \cline{2-8}
		 & LR & OC & FM & SV &TC&LI&OV \\  \hline
		GlobalTrack&0.434 & 0.552 & 0.839  & 0.857& 0.712 &0.840& 0.741\\
		SiamYOLO &0.598 & 0.706 &0.853 & 0.855& 0.753&0.850&0.778\\
		Unicorn&  0.716 &0.776 &  0.889 & 0.902& 0.774&0.884&0.785\\
		HCAT &0.180& 0.335 & 0.502& 0.559 & 0.174 &0.230&0.150\\
        SimTrack&0.290 & 0.460 &0.545  &0.581& 0.200&0.220&0.145\\
		ROMTrack &0.207 &0.335 & 0.581  &0.539&0.412 &0.230&0.347\\
        SMAT&0.158&0.302&0.499& 0.544 &0.162 &0.204&0.157\\
        HIT&0.309 &0.492&  0.694& 0.756&0.311 & 0.469&0.172\\   
        ODTrack&0.328 & 0.473& 0.769 &0.887 &0.407 &0.629&0.295\\
         MixFormer&0.344 &0.492&0.602 & 0.612& 0.434&0.266&0.402\\
          SiamDFF&\bf{0.836} & \bf{0.861}&\bf{0.910}&  \bf{0.913}& \bf{0.831}& \bf{0.930}&\bf{0.820}\\
        \hline
	\end{tabular}
	\label{nprecision}
\end{table}
To conduct a more comprehensive comparison of different trackers, we evaluate their performance in various attributes to comprehensively analyze the overall tracking capability of each tracker. These attributes include out-of-view (OV), occlusion (OC), fast motion (FM), scale variation (SV), low illumination (LI), thermal crossover (TC), and low resolution (LR). To better evaluate the performance of different trackers for small targets, we modify the scale definition. Specifically, we redefined the small-scale size from the original threshold of 400 to 80 which follows the definition in \cite{2022TIMFang,2023ACMMM_DANet,fang2024online}. Anti-UAV \cite{2023CVPRAntiUAVDatasets} datasets include attribute annotations for each video, and for our datasets, we meticulously labeled each frame of the sequence based on the description of each attribute.

The mSA, success, precision, and nprecision of all trackers are presented in Tables \ref{msa}, \ref{success}, \ref{precision}, and \ref{nprecision}, respectively. It is clear that SiamDFF achieves the best overall performance across all challenging scenarios, particularly in LR, OC, and TC. This demonstrates that SiamDFF can effectively extract fine-gained IRUT features in complex backgrounds.

\section{Qualitative Results of the Proposed Method on the Remaining Five Sequences}
\begin{figure*}[!t]
\centering
\includegraphics[width=\linewidth, keepaspectratio]{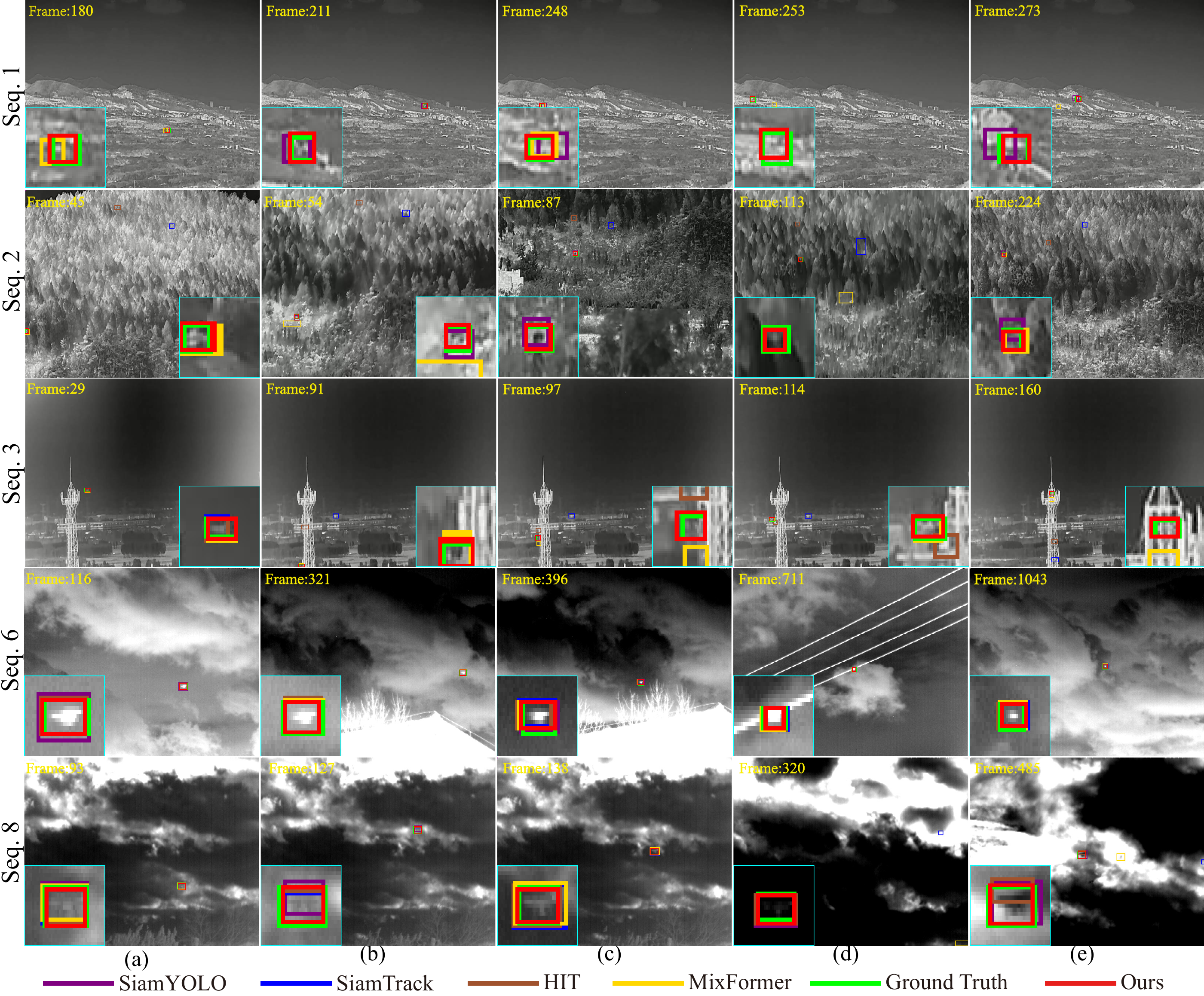}
\caption{Qualitative tracking results of five methods on five challenging sequences from Seq. 1-3, Seq. 6, and Seq. 8. Each row presents a sequence with five representative images. Close-ups are given in the left-bottom for better visualization.}
\label{tracking_results_all_methods-2}
\end{figure*}
We present the remaining qualitative results for Seq. 1-3, Seq. 6, and Seq. 8 in Fig. \ref{tracking_results_all_methods-2}. In Seq. 1 and Seq. 2, the target is completely immersed in the forest background, making it challenging for local search-based methods like SiamTrack and HIT to successfully re-track the IRUT when tracking failures occur. In Seq. 3(b), where the target appears at the edge of the image, only our method accurately estimates the target's shape. In Seq. 8(e), MixFormer incorrectly tracks the distractor, while HIT can only locate part of the target, highlighting their inability to extract discriminative features effectively. Although SiamYOLO manages to locate the target to some degree, it struggles to accurately estimate the target's size. In contrast, our method consistently outperforms other baseline methods in terms of location accuracy and scale estimation, clearly demonstrating the superiority of SiamDFF in real-world anti-UAV scenarios.

\section{Selection of the specified threshold $T$ in the module STEN}
In this section, we try to set the threshold $T$ as a learnable parameter. As shown in Table \ref{Threshold_T_Selection}, the final estimated threshold value $T$ is 0.6426, which is relatively close to the manually selected value of 0.7. In addition, we can observe in Table \ref{Threshold_T_Selection} that the tracking method with a learnable threshold parameter achieves performance very close to that of the method using a manually selected optimal threshold. This indicates that the threshold $T$ can be set as a learnable parameter.

\begin{table}[t!]
	\centering
	\caption{Impact of Threshold $T$ Selection on Algorithm Performance}
	\setlength\tabcolsep{2.5pt} 
	\fontsize{8.8}{10}\selectfont  
	\begin{tabular}{p{9em}<{\centering}|c|c|c|c} \hline
    Threshold $T$           &  mSA   & Success  & Precision  &  NPrecision   \\ \hline
 \makecell{0.7\\(Manual selection)}     & 0.686 & 0.682    & 0.877      &   0.832    \\  \hline
\makecell{0.6426\\(Learned parameter) } & 0.691 & 0.686     & 0.907      & 0.861  \\  \hline
	\end{tabular}
	\label{Threshold_T_Selection}
\end{table}

\section{Computational Complexity Comparison Between Our Method and the Standard Distillation Method}
\begin{table}[!t]
\centering
\caption{Computational complexity comparison between our method and the standard distillation method }
\fontsize{7}{9}\selectfont  
\vspace{-0.1em}
\setlength\tabcolsep{0.5em} 
\renewcommand{\arraystretch}{1.3}
\begin{tabular}{c|c|c|c|c|c|c} \hline
\multirow{2}{*}{Method}  &  \multicolumn{3}{c|}{Standard Distillation Method} & \multicolumn{3}{c}{Ours} \\  \cline{2-7}
                                  &  Backbone  & \makecell{Params\\ (M)} & \makecell{FLOPs\\ (G)} &  Backbone  &  \makecell{Params\\ (M)}  & \makecell{FLOPs\\ (G)}  \\  \hline  
Teacher                           & ResNet50 \cite{He_2016_CVPR}  & 25.5    & 12.30 & ConvNeXt \cite{2022CVPRConvNeXtLiu} & 25.94   & 4.19  \\   \hline
Student                           & ResNet18 \cite{He_2016_CVPR}  & 11.18   & 5.35  & ResNet18 \cite{He_2016_CVPR} & 11.18   & 5.35  \\  \hline
\makecell{Added\\ module}                     &  -          & -       & -     & TCAKD    & 37.11   & 35.34  \\  \hline
\makecell{Total\\ complexity}                & -            & 36.68   & 17.65 & -         & 74.23  &  44.88 \\  \hline
\end{tabular}
\label{Computational complexity}
\end{table}

In this section, we present a comparison of the computational complexity between our method and the standard distillation approach. Here, we mainly consider feature-based knowledge distillation methods built upon backbone networks. As illustrated in Table \ref{Computational complexity}, it can be seen that the proposed method has a certain increase in computational complexity compared to standard distillation method, mainly due to the introduction of a new attention module. However, this does not affect the inference efficiency of our algorithm.

\section{Additional Experimental Results on Public UAV410 Dataset}

\begin{table}[t!]
	\centering
\caption{Quantitative Comparisons of the Proposed Method and Other Methods on the UAV410 Dataset. The best three results are shown in \textcolor{red}{red}, \textcolor{green}{green}, and \textcolor{blue}{blue} fonts, respectively.}
	\setlength\tabcolsep{2.5pt} 
	\fontsize{8.8}{10}\selectfont  
	\begin{tabular}{p{5em}<{\centering}|c|c|c|c} \hline
Method       &  mSA $\uparrow$   & Success $\uparrow$  & Precision $\uparrow$  &  NPrecision $\uparrow$   \\ \hline
GlobalTrack  & \color{blue}{0.601} & \color{blue}{0.585} & \color{blue}{0.795} & \color{blue}{0.745} \\  \hline
Unicorn     & \color{green}{0.667} & \color{green}{0.650} & \color{green}{0.886} & \color{green}{0.809} \\  \hline
HCAT        & 0.137 & 0.133      & 0.167      & 0.167 \\  \hline
SimTrack     & 0.154 & 0.149     & 0.201     & 0.197 \\  \hline
ROMTrack     & 0.309 & 0.301    & 0.405      & 0.385 \\  \hline
SMAT        & 0.141 & 0.137     & 0.185     & 0.176 \\  \hline
HIT         & 0.221 & 0.215     & 0.290      & 0.285 \\  \hline
ODtrack     & 0.317 & 0.309     & 0.416      & 0.402 \\  \hline
MixFormer   & 0.322  & 0.313    & 0.421      & 0.396  \\  \hline
Ours        &\color{red}{0.721}  & \color{red}{0.703} & \color{red}{0.890}      & \color{red}{0.846}  \\  \hline
	\end{tabular}
	\label{additional_results_UAV410}
\end{table}

\begin{figure}[!t]
\centering
\includegraphics[width=0.96\linewidth, keepaspectratio]{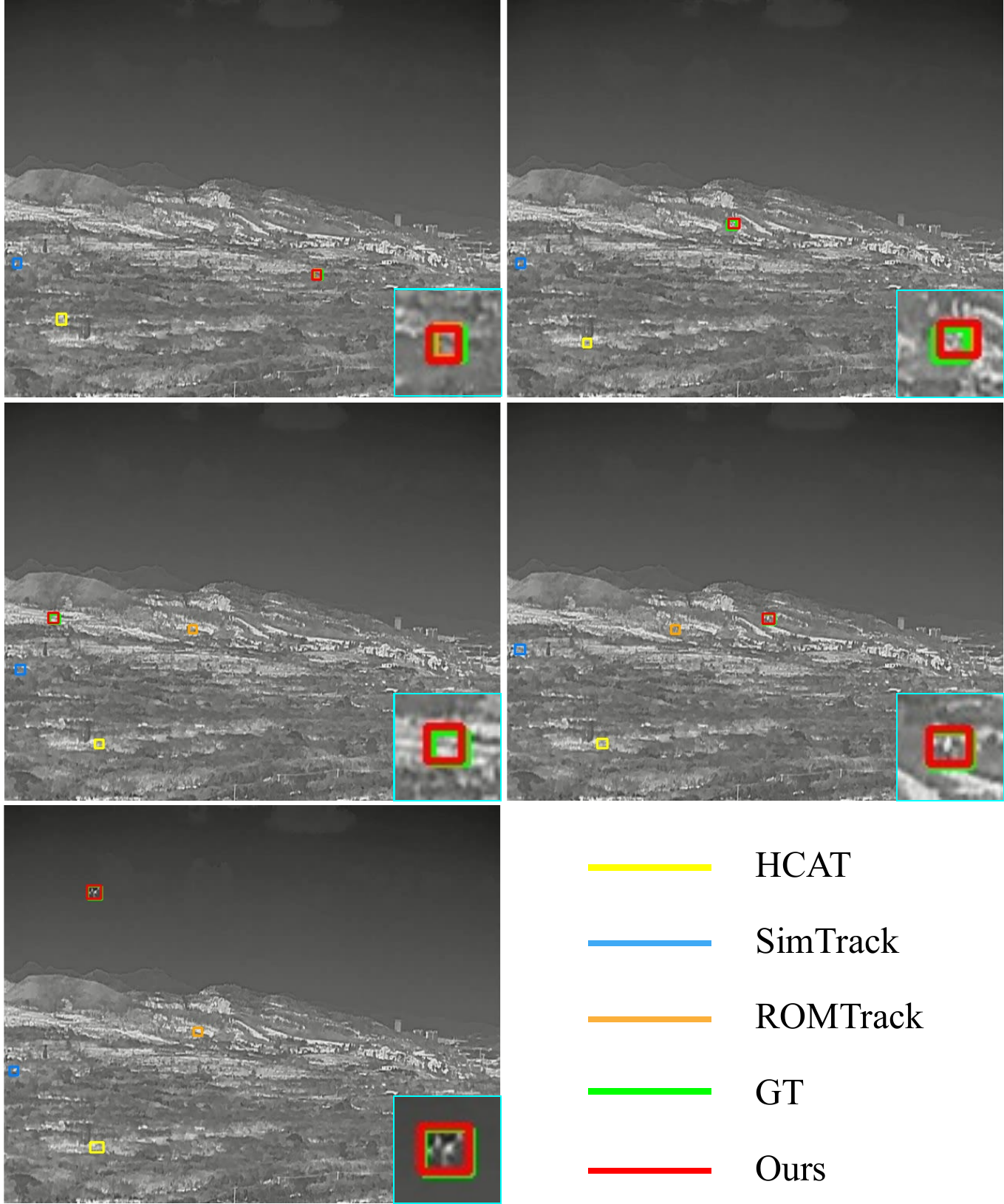}
\caption{Tracking results of our method and other methods on the UAV410 dataset.}
\label{track_result_uav410}
\end{figure}

In this section, we present further experimental results on UAV410 dataset \cite{2021TMMJiang} to validate the generalization and effectiveness of the proposed method. As shown in Table \ref{additional_results_UAV410}, we can see that our method achieves better performance compared with other methods. As presented in Fig. \ref{track_result_uav410}, we also shows the infrared target tracking results of HCAT, SimTrack, ROMTrack,  and our method from UAV410 dataset. We can observe that our method consistently achieves reliable tracking of UAV targets, whereas all other approaches fail to maintain target tracking.

\section{Minimum detectable target size and  fast-moving UAV detection}
\begin{table}[t!]
	\centering
\caption{Quantitative Comparisons of the Proposed Method and Other Methods for Fast-moving UAV Targets. The best three results are shown in \textbf{Bold}.}
	\setlength\tabcolsep{2.5pt} 
	\fontsize{10}{13}\selectfont  
	\begin{tabular}{p{5em}<{\centering}|c|c|c|c} \hline
Method       &  mSA $\uparrow$   & Success $\uparrow$  & Precision $\uparrow$  &  NPrecision $\uparrow$   \\ \hline
ROMTrack     & 0.234 & 0.228   & 0.300      & 0.277 \\  \hline
HIT         & 0.166 & 0.162     & 0.290      & 0.200 \\  \hline
SimTrack     & 0.168 & 0.164     & 0.215     & 0.197 \\  \hline
Ours        &\textbf{0.636}  & \textbf{0.620} & \textbf{0.794}      & \textbf{0.755}  \\  \hline
	\end{tabular}
	\label{fast_moving_uav}
\end{table}

\begin{figure}[!t]
\centering
\includegraphics[width=0.96\linewidth, keepaspectratio]{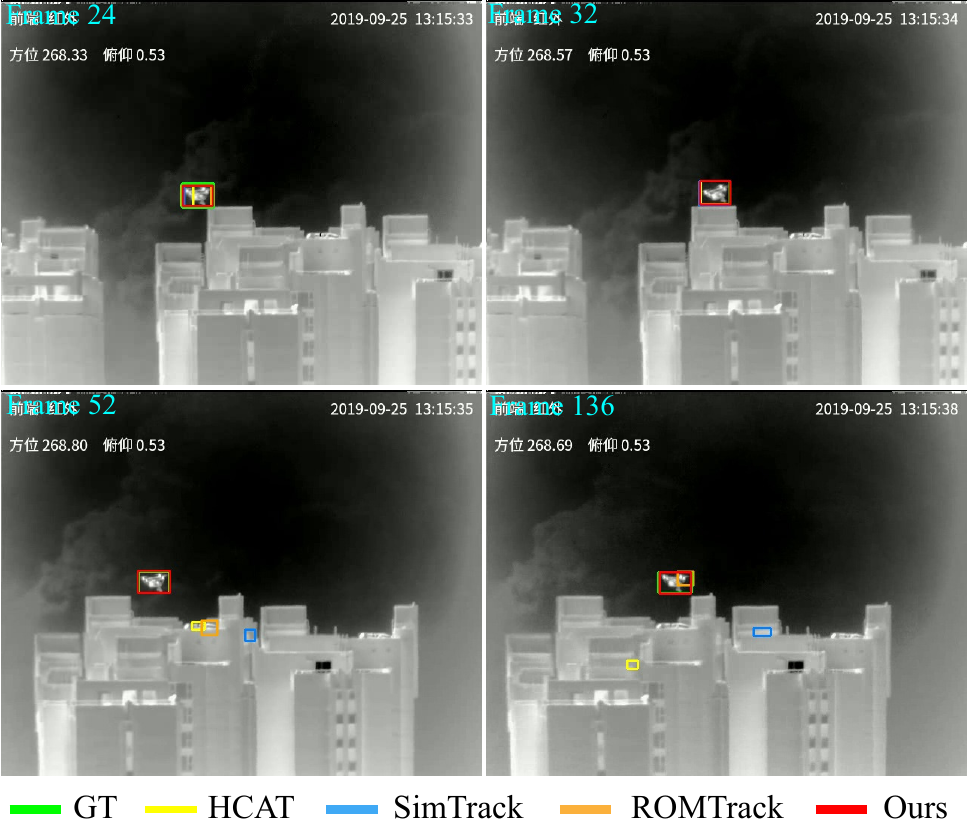}
\caption{Tracking results of our method and other methods on extremely fast-moving UAV dataset.}
\label{fast_moving_uav_track}
\end{figure}

In this section, we calculated that the minimum detectable target size of the proposed model is 5$\times5$ pixels. We extract 200 frames from the 1000 frames of Seq. 7, which exhibit significant background variation and target position changes. These selected frames are used to simulate scenarios of extremely fast-moving UAVs. We also present a comparison of the tracking performance between the proposed method and other approaches under the extremely fast-moving UAV scenario, as shown in Table \ref{fast_moving_uav}. In the fast-moving UAV scenario, the background of the target changes drastically, posing significant challenges for tracking algorithms. As shown in Table \ref{fast_moving_uav} and Fig. \ref{fast_moving_uav_track}, the proposed method still achieves excellent tracking performance on the fast-moving UAV target dataset. This is because the proposed method is a long-term robust tracker capable of handling situations where the target undergoes substantial positional changes across multiple frames. It can be observed from Fig.  \ref{fast_moving_uav_track} that existing tracking methods fail when the frame interval is large (e.g., from frame 32 to frame 52), whereas our method is still able to stably track the UAV target. This verifies the effectiveness of the proposed approach.

\ifCLASSOPTIONcaptionsoff
\newpage
\fi
\bibliography{reference_supp}

\end{document}